\pdfoutput=1

\documentclass[a4paper,12pt]{book}

\usepackage[utf8]{inputenc}
\usepackage[pagebackref,hyperindex=true,breaklinks=true]{hyperref} 
\usepackage{breakcites} 
\usepackage[pagebackref,hyperindex=true]{hyperref} 
\usepackage[ruled,vlined]{algorithm2e}
\usepackage{amsmath}
\usepackage{amsthm}

\usepackage{graphicx}
\usepackage{caption}
\usepackage{subcaption}

\theoremstyle{plain}
\newtheorem{lemma}{Lemma}
\newtheorem{proposition}{Proposition}

\theoremstyle{definition}
\newtheorem{definition}{Definition}[section]
\newtheorem{condition}{Condition}[section]

\theoremstyle{remark}

\usepackage{xcolor}

\definecolor{linkcol}{rgb}{0,0,0.4}
\definecolor{citecol}{rgb}{0.5,0,0}

\renewcommand*{\backref}[1]{}
\renewcommand*{\backrefalt}[4]{%
\ifcase #1 %
(Not cited.)%
\or
(Cited on page~#2.)%
\else
(Cited on pages~#2.)%
\fi}

\usepackage[ngerman,english]{babel}
\usepackage[T1]{fontenc}
\usepackage{amsfonts} 
\usepackage[font=footnotesize,labelfont=bf]{caption} 
\usepackage{charter}

\let\tmp\oddsidemargin
\let\oddsidemargin\evensidemargin
\let\evensidemargin\tmp
\reversemarginpar

\usepackage{todonotes}

\newcommand{\EE}{\mathbb{E}}        
\newcommand{\RR}{\mathbb{R}}        

\def\addsymbol #1: #2#3{$#1$ \> \parbox{115mm}{#2 \dotfill \pageref{#3}}\\}
\def\newnot#1{\label{#1}}

\hypersetup
{
bookmarksopen=true,
pdftitle=Implementing spectral methods for hidden Markov models with real-valued emissions,
pdfauthor=Carl Mattfeld,
pdfsubject=ShortTitle, 
pdfmenubar=true, 
pdfhighlight=/O, 
colorlinks=true, 
pdfpagemode=UseNone, 
pdfpagelayout=SinglePage, 
pdffitwindow=true, 
linkcolor=linkcol, 
citecolor=citecol, 
urlcolor=linkcol 
}

 \newcommand{\HRule}{\rule{\linewidth}{0.5mm}}

\begin{document}

\frontmatter

\begin{titlepage}
\begin{center}


\textsc{\LARGE ETH Z\"urich}\\[0.5cm]
\textsc{\Large Department of Physics}\\[1.5cm]

\textsc{\Large Master Thesis}\\[0.5cm]

\HRule \\[0.4cm]
{ \huge \bfseries Implementing spectral methods for hidden Markov models with real-valued emissions \\[0.4cm] }

\HRule \\[1.3cm]

\begin{flushleft}
\Large \emph{Author:}\\
\Large Carl Mattfeld$^{1,2}$ \\[1.3cm]
\end{flushleft}

\begin{flushleft} \large
\emph{Supervisors:} \\
Prof.~Dr. Chris Wiggins$^1$\\
Prof.~Dr. Didier Sornette$^2$\\[1cm]

\normalsize
$^1$ Columbia University, New York, NY 10027, USA\\
$^2$ ETH Z\"urich, 8092 Z\"urich, Switzerland

\end{flushleft}

\vfill

{\large \today}

\end{center}
\end{titlepage}

    \null\vspace{\stretch {1}}
        \begin{flushright}
                To my family.
        \end{flushright}
\vspace{\stretch{2}}\null

\newenvironment{acknowledgements}%
{\cleardoublepage\null\vfill\begin{center}
    \bfseries Acknowledgements\end{center}}%
    {\vfill\null}
\begin{acknowledgements}
I would like to thank Professor Chris Wiggins for supervising this thesis and challenging me with this project. I am grateful to have had this opportunity to join him as a visiting researcher at Columbia University.\\
I would like to thank Professor Didier Sornette for being available as my supervisor at ETH Zurich.\\
Furthermore, I would like to thank Professor Daniel Hsu for the valuable discussions. These discussions were immensely helpful for understanding his original work and getting important ideas, especially for the binning of real-valued emissions.\\
I am highly grateful for being sponsored by a stipend of the German National Academic Foundation (\textit{Studienstiftung des deutschen Volkes}). This stipend included financial sponsorship for my regular studies in Switzerland, a semester abroad in Hong Kong, and this research project in New York. I would like to thank Professor Dagmar Iber, my liaison professor with the Studienstiftung, and Marius Spiecker gen. D\"ohmann, my advisor with the Studienstiftung in Bonn.\\
Finally, I would like to thank my family and my friends for their support and being there for me.
\end{acknowledgements}

\newenvironment{abstract}%
		{\cleardoublepage\null\vfill\begin{center}
    \bfseries\abstractname\end{center}}%
    {\vfill\null}
        \begin{abstract}
				Hidden Markov models (HMMs) are widely used statistical models for modeling sequential data. The parameter estimation for HMMs from time series data is an important learning problem. The predominant methods for parameter estimation are based on local search heuristics, most notably the expectation–maximization (EM) algorithm. These methods are prone to local optima and oftentimes suffer from high computational and sample complexity. Recent years saw the emergence of spectral methods for the parameter estimation of HMMs, based on a method of moments approach. Two spectral learning algorithms as proposed by \cite{Hsu2012} and \cite{AHK2012} are assessed in this work. Using experiments with synthetic data, the algorithms are compared with each other. Furthermore, the spectral methods are compared to the Baum-Welch algorithm, a well-established method applying the EM algorithm to HMMs. The spectral algorithms are found to have a much more favorable computational and sample complexity. Even though the algorithms readily handle high dimensional observation spaces, instability issues are encountered in this regime. In view of learning from real-world experimental data, the representation of real-valued observations for the use in spectral methods is discussed, presenting possible methods to represent data for the use in the learning algorithms.
        \end{abstract}

\tableofcontents

\listoffigures

\mainmatter

\chapter{Introduction}
\label{cha:introduction}
Hidden Markov models (HMMs) are important statistical models used to describe real-world processes that produce sequential data. The underlying systems are assumed to be Markov processes, with the states of the processes not directly observable (\textit{hidden}). Only a sequence of emissions is observable. HMMs are used in several applications, including speech recognition \cite{Rabiner1989}, natural language processing (NLP) \cite{Manning1999}, and the analysis of biological processes, such as protein topology \cite{Krogh2001}. \\
The predominant methods for estimating HMM parameters rely on local search heuristics. The \textit{Baum-Welch (BW) algorithm} \cite{Baum1970} and related methods are widely used. They utilize the \textit{expectation-maximization (EM) algorithm} to find maximum likelihood parameter estimates (\cite{Dempster1977} and \cite{Bilmes1998}). Unfortunately, these methods are prone to local optima and slow convergence \cite{Redner1984}, and oftentimes suffer from high computational and sample complexity.\\
An alternative approach to the HMM learning problem is the \textit{method of moments}. This parameter estimation technique was first proposed by \cite{Pearson1894}. The underlying idea is to sample empirical moments from experimental data, and then to find model parameters that yield expected values equal to the sampled quantities. Efficient learning algorithms for HMMs using this approach were enabled by a spectral decomposition technique introduced by \cite{Chang1996}. These algorithms can also be referred to as \textit{spectral methods}. In contrast to the iterative approach used by the local search heuristics, spectral methods are not susceptible to local optima. Especially when learning parameter estimates from large observation spaces, this is an important advantage.\\
In this work at hand, the use of spectral methods for learning hidden Markov models is investigated. Spectral algorithms by \cite{Hsu2012} and \cite{AHK2012} are introduced and characterized using synthetic experimental data.\\
Considering one of the most basic real-world settings, a measurement setup is likely to produce experimental data discretized in time with real-valued data. Assuming a hidden Markov model is to be used to describe the underlying process, it is important to find an appropriate way to prepare the experimental data for the use in the learning algorithms. Besides characterizing the spectral algorithms, this work will therefore also discuss how the algorithms could be used to obtain parameter estimates from sequences of real-valued observations.\\
For further reading on learning problems and machine learning in general, the book \cite{Bishop2006} provides a comprehensive treatment and can be recommended.

\section{Outline}
In this introductory chapter, an overview over the state of research is given, followed by an introduction to hidden Markov models.\\
The second chapter presents the two spectral learning algorithms by \cite{Hsu2012} and \cite{AHK2012}. For the latter algorithm, a detailed derivation will be given. The last section of the second chapter addresses how real-valued emissions can be used as an input for the learning algorithms.\\
The third chapter presents and discusses the results of the experiments conducted in this work. First, the two spectral algorithms will be compared with each other. In a second part, the algorithm by \cite{AHK2012} will be compared with the Baum-Welch algorithm. In the last section, an experimental analysis of the use of real-valued emissions as the input data for spectral algorithms is given.\\
Chapter four provides the conclusion of this work and gives an outlook on possible future work.\\

\section{State of research}
Spectral methods for learning hidden Markov models were only established rather recently. Even though the underlying method of moments is a rather old concept, a major issue was that the estimation of high-order moments resulted in a high computational and sample complexity. \\
A spectral decomposition technique proposed by \cite{Chang1996} can be considered a milestone in enabling efficient spectral learning algorithms for HMMs. The method recovers model parameters using moments only up to third-order. The work also showed that second-order moments are not sufficient to fully identify the parameters of Markov models.\\
Based on this spectral decomposition, \cite{Mossel2005} proposed an efficient learning algorithm for phylogenic trees and HMMs. Compared to this algorithm by \cite{Mossel2005}, the two algorithms that are under consideration in the work at hand share a similar rank condition on the emission and transition matrix of a HMM, and use similar empirical quantities computed from the observations for learning. \cite{Mossel2005} require the emission and transition matrix to have the same dimension.\\
The algorithm by \cite{Hsu2012}, which is the subject of this work, relaxed this assumption, allowing a larger observation space. The emission and transition matrix of the HMM are still assumed to have full rank. In favor of a higher sample-efficiency, this algorithm does not yield empirical estimates for the emission and transition matrices though. Instead, a so-called observable representation of probabilistic quantities of a HMM is given. The work by \cite{Hsu2012} was initially published to the \textit{arXiv.org} repository in 2009 and resulted in a work by \cite{Siddiqi2009}.\\
\cite{Siddiqi2009} relaxed the assumption for the transition matrix to have full rank, resulting in an extended application of the algorithm by \cite{Hsu2012} to the case of so-called \textit{Reduced-Rank Hidden Markov Models} (RR-HMMs). In RR-HMMs, the dynamics are assumed to be evolving in a subspace smaller than the dimension of the transition matrix.\\
The second algorithm that is the subject of the work at hand was proposed by \cite{AHK2012}. The algorithm follows the same approach as the algorithms mentioned above, and assumes the same full-rank condition. The learning algorithm for HMMs is proposed as a special case of a general learning algorithm for multi-view mixture models.\\
The works by \cite{Mossel2005}, \cite{AHK2012} and \cite{Hsu2012} (plus the related algorithm by \cite{Siddiqi2009}) utilize some variant of a diagonalization of a collection of similar matrices obtained from a tensor. The work by \cite{AGHKT2012} addresses the issue of separation among eigenvalues when calculating model parameters from the diagonalization, which can lead to instability. \cite{AGHKT2012} proposes a tensor power method allowing a more stable calculation of model parameters from an experimentally obtained tensor quantity. This theoretical work addresses latent variable models in general, listing hidden Markov models as a possible application. An explicit learning algorithm for HMMs is not formulated.

\section{Notation}

For two vectors $\vec{v}$ and $\vec{w}$, we denote the standard inner product by $\left\langle \vec{v}, \vec{w}\right\rangle = \vec{v}^\top \vec{w}$. \\
Let $\Delta^{n-1}:= \left\{ (p_1,p_2,\ldots,p_n) \in \RR^n \, | \, p_i \geq 0 \, \forall \, i, \sum_{i=1}^n p_i =1 \right\}$ denote the probability simplex in $\RR^n$.\\
For a matrix $A \in \RR^{m \times n}$, we let $\|A\|_F=\sqrt{\sum_{i=1}^m\sum_{j=1}^n |a_{ij}|^2}$ denote its Frobenius norm and $A^+$ its Moore–Penrose pseudoinverse. \\
Let $\vec{u} \in \RR^n$ and $B \in \RR^{m \times n}$ be the analytical quantities of a vector and a matrix. We then denote their empirical estimates from experimental data by $\hat{\vec{u}} \in \RR^n$ and $\hat{B} \in \RR^{m \times n}$, respectively.

\section{Hidden Markov models}
\label{sec:hmmintro}
Hidden Markov models were developed by Leonard E. Baum and coworkers in \cite{Baum1966} and subsequent works. In this section, the mathematical foundations are introduced. In a first step, the concept of a Markov process is presented. The introduction of latent variables then leads to the hidden Markov model. Model parameters are introduced as probabilistic measures, leading to the formulation of the learning problem. The mathematical presentation in this section follows \cite{Bishop2006}.

\subsection{Markov process}
Let $p(\cdot)$ denote a probability distribution function. Let there be a sequence of $L$ observations, with $s_t$ denoting the observation at time $t \in  \left\{1,...,L\right\}$. The observation can be a vector or scalar.
The joint distribution of a sequence of $L$ observations is given by
\begin{equation}
	p(s_1,\ldots,s_L) = p(s_1) \prod_{t=2}^L p(s_t|s_1,\ldots,s_{t-1}).
\end{equation}
Assuming the Markov property, the current observation in a sequence only depends on the previous one. This is also known as a \textit{Markov chain}. For a sequence of observations, the joint distribution can therefore be written as
\begin{equation}
	p(s_1,\ldots,s_L) = p(s_1) \prod_{t=2}^L p(s_t|s_{t-1}).
\end{equation}
One can verify that, given previous observations up to time $t$, this yields the following conditional distribution for an observation $s_t$:
\begin{equation}
	p(s_t|s_1,\ldots,s_{t-1}) = p(s_t|s_{t-1}).
\end{equation}
In a hidden Markov model, we now assume an underlying Markov process that cannot be observed. While the states are hidden, they emit an output which is observable. In the model, each state has a corresponding probability distribution over the possible emissions. The hidden states are interpreted as discrete \textit{latent variables} which are not directly observable. These latent variables are then assumed to form a Markov chain. The possible emissions can be either discrete or continuous. \\

\subsection{Hidden states as latent variables}
\label{sec:hiddenstates}
\begin{figure}
	\centering\includegraphics[scale=0.5]{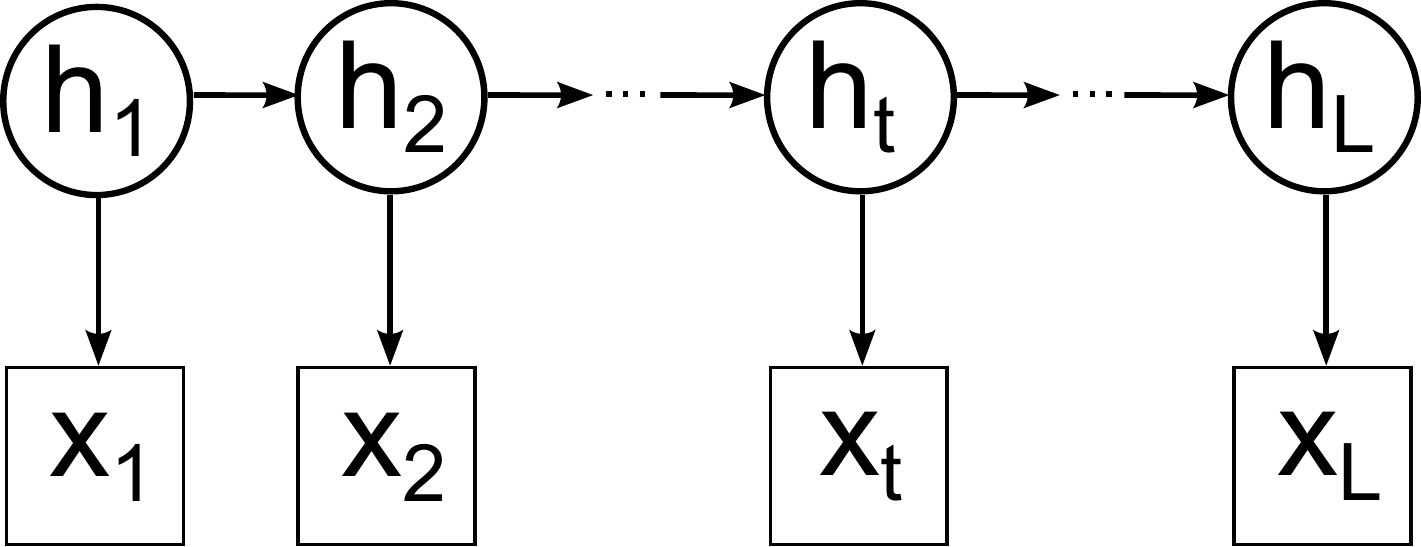}
	\caption{A graphical structure representing a hidden Markov model: a sequence of discrete hidden states $\left\{h_t\right\}_{t=1,...,L}$ with the corresponding emissions $\left\{x_t\right\}_{t=1,...,L}$.}
	\label{fig:HMMgraph}
\end{figure}
We assume that the underlying states form an unobservable Markov chain. We denote a hidden state at time $t$ by $h_t$. The possible discrete or continuous emissions by this state are denoted by $x_t$. The resulting graphical structure is depicted in Figure \ref{fig:HMMgraph}.\\
This setup now allows the parameterization of a hidden Markov model only by the transition probability between states, the probability distribution over possible emissions by each state, and the initial distribution of states. \\

The joint distribution for a hidden Markov model is given by
\begin{equation}
	p(x_1,...,x_L,h_1,...,h_L) = p(h_1) \left[ \prod_{t=2}^{L} p(h_{t+1}|h_{t}) \right] \prod_{t=1}^L p(x_t|h_t).
\end{equation}

\subsection{State transitions}
\label{sec:statetrans}

Let there be $k$ possible hidden states. Since the latent variables in a HMM are discrete, we will denote the set of hidden states by $\left[ k \right] = \left\{ 1,...,k \right\}$ \newnot{symbol:setofstates}. In a HMM, the transition probabilities between two states do not evolve with time. Therefore we can introduce a \textit{state transition probability matrix }. Let $T \in \mathbb{R}^{k \times k}$ \newnot{symbol:transitionmatrix} be such a matrix with
\begin{equation}
\label{eq:transelements}
	T_{ij} = \Pr \left[h_{t+1}=i | h_t=j\right], \quad i,j \in \left[ k \right].
\end{equation}
$T_{ij}$ is the probability for the latent variable $h_{t+1}$ to be in the state $i \in \left[ k \right]$ at time $t+1$ after having been in state $j \in \left[ k \right]$ at time $t$. The elements of the matrix are probabilities, and therefore $T_{ij} \in [0,1]$ with $\sum_{i=1}^k T_{ij} = 1, \, \, \forall j \in \left[ k \right]$. This results in $k(k-1)$ independent parameters for the transition matrix. \\

For the initial state $h_1$, the \textit{initial state distribution} is denoted by the vector $\vec{\pi} \in \mathbb{R}^k$ \newnot{symbol:initstatedist}, with 
\begin{equation}
\label{eq:initialvector}
	\pi_j = \Pr \left[ h_1=j  \right]. 
\end{equation}

\subsection{Emissions}
\label{sec:emissions}
The emissions of a hidden Markov model can be parameterized by an \textit{emission probability matrix}. The simplest case is for \textit{categorical} emissions. Here the emission probability matrix can be intuitively understood as a probability table.\\
Let there be $d$ discrete observable emissions, with $d \geq k$. The set of discrete emissions can be written as $\left[ d \right] = \left\{ 1,...,d \right\}$ \newnot{symbol:setofobservations}. The emission $x_t=i, i \in \left[ d \right]$ was emitted by the hidden state $h_t=j, j \in \left[ k \right]$ with the probability $\Pr \left[ x_t=i|h_t=j \right]$. This now allows the introduction of the emission probability matrix $O \in \mathbb{R}^{d \times k}$ \newnot{symbol:observationmatrix}, with
\begin{equation}
\label{eq:obselements}
	O_{ij} = \Pr \left[ x_t=i | h_t=j  \right], \quad i \in \left[ d \right], j \in \left[ k \right].
\end{equation}
Another case are \textit{vector-valued} emissions. Here an emission at time $t$ is represented by a random vector $\vec{x}_t$ with values in $\RR^d$. In this case we can introduce the emission matrix as
\[
	O = \left[ \vec{o}_1 | \vec{o}_2 | \ldots | \vec{o}_k \right] \in \mathbb{R}^{d \times k}.
\]
The columns of the matrix denote the conditional means of the emissions $\vec{x}_t$ at time $t$, given the corresponding discrete hidden state $h_t$:
\[
	\mathbb{E} \left[ \vec{x}_t | h_t=j \right] = O \vec{e}_j = \vec{o}_j, \quad j \in \left[ k \right],
\]
where $\vec{e}_j$ is the $j$-th vector of the $d$-dimensional standard basis. The conditional mean of an emission could for example represent the mean of a multivariate Gaussian.\\
The words \textit{emission} and \textit{observation} are often used interchangeably. In this work, we understand that in the generative process, a hidden state emits an output which can be observed. This output can be interpreted both as the emission of the hidden state, or the observation of this emission. Especially in the experimental context, it can be considered more appropriate to talk about an observation, rather than an emission.\\
Both terms will be used in this work. The term \textit{observation} will be preferred in the context of experimental data, while the term \textit{emission} will be used when referring to the generative process.

\subsection{Example}

\begin{figure}
	\centering\includegraphics[scale=0.8]{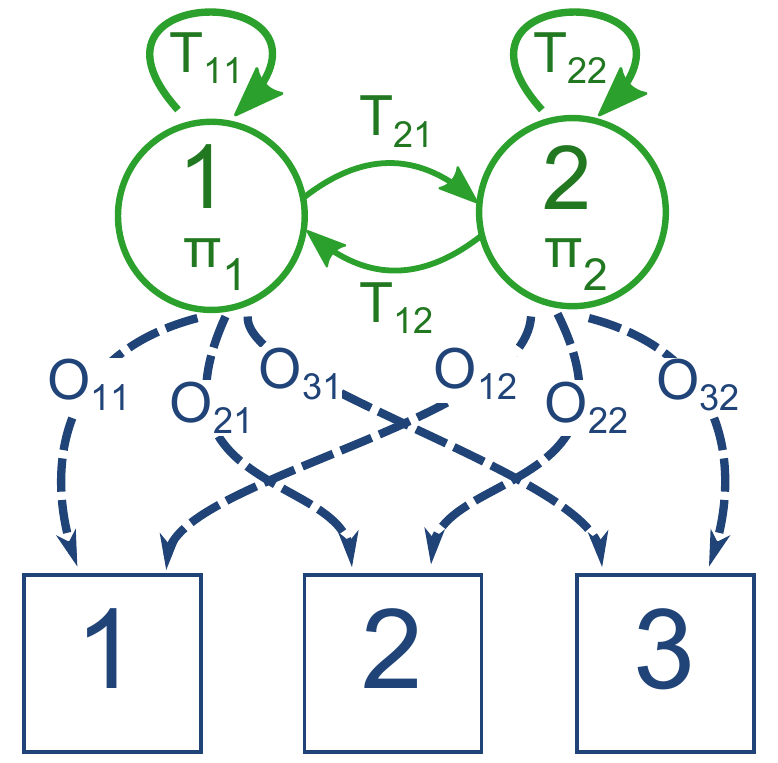}
	\caption{A two-state HMM (states 1 and 2) with three possible emissions (1, 2 and 3). The initial state distribution is indicated inside the circles of the corresponding states ($\pi_1$ and $\pi_2$). Solid arrows indicate transitions between hidden states, dashed arrows indicate emissions of the possible observations.}
	\label{fig:HMMex}
\end{figure}

As an example, we now consider a two-state hidden Markov model as depicted in Figure \ref{fig:HMMex}. The underlying system can be in the hidden states $1$ or $2$. At a given timestep, the system then emits one of the three observations $1$,$2$ or $3$, depending on which hidden state it is in. Using equations \eqref{eq:transelements}, \eqref{eq:initialvector} and \eqref{eq:obselements}, the model can then be parameterized in the following way: let $T \in \RR^{2 \times 2}$, $\vec{{\pi}} \in \RR^2$ and $O \in \RR^{3 \times 2}$, then
\begin{equation}
T =
 \begin{pmatrix}
  T_{11} & T_{12} \\
  T_{21} & T_{22}
 \end{pmatrix}
 = 
	\begin{pmatrix}
  9/10 & 3/10 \\
  1/10 & 7/10
 \end{pmatrix},
\end{equation}
\begin{equation}
	\vec{\pi} = 
	\begin{pmatrix}
  \pi_1 \\
  \pi_2
 \end{pmatrix}
 = 
 \begin{pmatrix}
  4/5 \\
  1/5
 \end{pmatrix}
\end{equation}
and
\begin{equation}
O =
 \begin{pmatrix}
  O_{11} & O_{12}\\
  O_{21} & O_{22} \\
  O_{31} & O_{32}
 \end{pmatrix}
 =
 \begin{pmatrix}
  1/4 & 8/10 \\
  1/2 & 1/10 \\
	1/4 & 1/10
 \end{pmatrix}.
\end{equation}
The numerical values given are exemplary values. A HMM parameterized by these matrices will be one of the models used in our analysis of the learning algorithms in Chapter \ref{cha:results_and_discussion}.

\subsection{Learning problem}
The learning problem for hidden Markov models is generally an instance of \textit{unsupervised learning}. In unsupervised learning, the experimental data is unlabeled, i.e. the desired output is not known. \footnote{For further information, refer to \cite{Bishop2006}.}\\
A HMM can be characterized by two model parameters, the number of states $k$ and the number (or dimension) of emissions $d$, and three probability measures, $T$, $O$ and $\vec{\pi}$. \footnote{We only consider the case of a homogeneous model where the parameters $T$,$O$,$\vec{\pi}$ represent all corresponding conditional probabilities for the respective states and emissions.} In a more compact notation, the probability measures are often grouped as a complete parameter set of the model:
\newnot{symbol:lambdaparameter}
\[
	\lambda = \left(T,O,\vec{\pi}\right).
\]
Let $p(x_1,\ldots,x_L|\lambda)$ be the probability of observing the sequence $\left(x_1,\ldots,x_L\right)$, provided the model parameters $\lambda$. For a hidden Markov model to be useful in a real-world application, \cite{Rabiner1989} proposed the following three basic problems of interest:
\begin{enumerate}
	\item Given a sequence of observations $\left(x_1,\ldots,x_L\right)$ and a model $\lambda = \left(T,O,\vec{\pi}\right)$, how can $p(x_1,\ldots,x_L|\lambda)$ be computed efficiently?
	\item Given a sequence of  observations $\left(x_1,\ldots,x_L\right)$ and the model $\lambda$, how do we choose a (scientifically) meaningful or optimal corresponding sequence of states $\left(h_1,\ldots,h_L\right)$?
	\item How can the model parameters $\lambda = \left(T,O,\vec{\pi}\right)$ be chosen such that they best account for the observed signal (i.e. maximize $p(x_1,\ldots,x_L|\lambda)$)?
\end{enumerate}
This work will address the last of the three basic problems. In our setting, we will assume we are given a series of observations. These observations will then be used to learn the model parameters $\lambda$ of the hidden Markov model. \\

\chapter{Theory}
\label{cha:theory}
This chapter will introduce two spectral algorithms. The algorithm by \cite{Hsu2012} will be presented in Section \ref{sec:HKZalg}. The presentation will be kept rather brief, since a comprehensive derivation is given in the original work. Furthermore, we will refrain from introducing the concept of observable representations associated with this algorithm, since it will not be relevant for our considerations.\\
The algorithm by \cite{AHK2012} will be presented in Section \ref{sec:AHK}. The original work does not explicitly formulate a learning algorithm for hidden Markov models. Therefore, we will give a comprehensive derivation of the mathematical considerations leading to the formulation of the spectral learning algorithm for HMMs. \\
Section \ref{sec:real_valued_section} will discuss real-valued emissions.


\section{Spectral algorithm by Hsu et al.}
\label{sec:HKZalg}
Here, we will introduce the spectral learning algorithm proposed by \cite{Hsu2012} for HMMs with discrete emissions. The algorithm in its basic form does not explicitly learn parameter estimates for the transition and emission probability matrices. It only learns a so-called observable representation allowing the estimation of the probability of observing a given sequence, and the approximation of the conditional probability of a future observation. \\
\cite{Hsu2012} points out a method to extend the algorithm to allow an estimation of the state transition probability matrix and emission probability matrix. Therefore, we will introduce the algorithm disregarding its observable representation functionality, and include the method to recover estimates for the state transition probability and emission probability matrices. It has to be noted, however, that the authors in \cite{Hsu2012} explicitly state that this additional estimation technique is to be considered generally unstable.\\
Let our HMM be parameterized as in Section \ref{sec:hmmintro}, with our observations being discrete. We then have $k$ hidden states denoted by $\left[ k \right] = \left\{ 1,\ldots,k \right\}$ and $d$ discrete observations denoted by $\left[ d \right] = \left\{ 1,\ldots,d \right\}$ with $d \geq k$. Our state transition probability matrix $T \in \RR^{k \times k}$ is then given by \eqref{eq:transelements}, the emission probability matrix $O \in \RR^{d \times k}$ by \eqref{eq:obselements} and the initial state distribution $\vec{\pi} \in \RR^k$ by \eqref{eq:initialvector}.\\
We assume the following condition on the HMM: 
\begin{condition}
\label{condition1HKZ}
	(HMM rank condition). $\pi_j > 0$ for all $j \in \left[k\right]$, and $\mathrm{rank} (O) = \mathrm{rank} (T) = k $.
\end{condition}
Furthermore, we define the following vector and matrix quantities:
\begin{definition}
\label{def:HKZ1}
Let $P_1 \in \RR^d$, $P_{2,1} \in \RR^{d \times d}$, $P_{3,1} \in \RR^{d \times d}$ and $P_{3,r,1} \in \RR^{d \times d}$ with
\begin{align*}
	&\left[P_1\right]_i = \Pr \left[x_1 = i \right], \\
	&\left[P_{2,1}\right]_{ij} = \Pr \left[x_2 = i, x_1 = j \right], \\
	&\left[P_{3,1}\right]_{ij} = \Pr \left[x_3 = i, x_1 = j \right], \text{and} \\
	&\left[P_{3,r,1}\right]_{ij} = \Pr \left[x_3 = i, x_2 = r, x_1 = j \right] \, \, \, \forall r \in \left[ d \right].
	\end{align*}
\end{definition}
These quantities are marginal probabilities of singletons, pairs and triples of observations. These quantities will later be sampled from the experimental data. \\
The algorithm utilizes a singular value decomposition of the empirically sampled estimation matrix of $P_{2,1}$. Let the matrix of the top $k$ left singular vectors of $P_{2,1}$ be denoted by $U \in \RR^{d \times k}$. We assume that the matrix obeys the following condition:
\begin{condition}
\label{condition2HKZ}
	(Invertibility condition). $U^\top O$ is invertible.
\end{condition}

\subsection{Algorithm learnHKZ}
The algorithm stated in \cite{Hsu2012} does not explicitly learn the state transition probability matrix and emission probability matrix of a HMM. However, the work also proposes a technique to learn these parameters. The resulting spectral learning algorithm is given by \\

\begin{algorithm}[H]
\label{alg:learnHKZ}
\SetAlgoLined
 \KwData{$N$ triples of observations, $k$ - number of states , $d$ - number of observations}
 \KwResult{Hidden Markov model parameterized by $\hat{O}$, $\hat{T}$ and $\hat{\vec{\pi}}$}

\begin{enumerate}
	\item Form the empirical estimates $\hat{P}_1 \in \RR^d$, $\hat{P}_{2,1} \in \RR^{d \times d}$, $\hat{P}_{3,1} \in \RR^{d \times d}$ and $\hat{P}_{3,r,1} \in \RR^{d \times d} \, \, \, \forall r \in \left[ d \right]$ using the $N$ triples of observations.
	\item For the $k$ largest singular values of $\hat{P}_{2,1}$, find the matrix $\hat{U} \in \RR^{d \times k}$ of left singular vectors.
	\item For all $r \in \left[ d \right]$, find the eigenvalues $\lambda_{r,1},\ldots,\lambda_{r,k}$ of 
		\[
			\left( \hat{U}^\top \hat{P}_{3,r,1} \right) \left( \hat{U}^\top \hat{P}_{3,1} \right)^+
		\]
	\item Return the empirical estimate of the matrix $\hat{O}$ given by:
		\[
			\hat{O} = 
			\begin{pmatrix}
				\lambda_{1,1} & \lambda_{1,2} & \cdots & \lambda_{1,k} \\
				\lambda_{2,1} & \lambda_{2,2} & \cdots & \lambda_{2,k} \\
				\vdots & \vdots & \ddots & \vdots \\
				\lambda_{d,1} & \lambda_{d,2} & \cdots & \lambda_{d,k}
			\end{pmatrix}
		\]
	\item Return $\hat{\vec{\pi}} = \hat{O}^+ \hat{P}_1$.
	\item Return $\hat{T} = \hat{O}^+ \hat{P}_{2,1} ( \hat{O}^+ )^\top \mathrm{diag} ( \hat{\vec{\pi}} )^{-1}$.
 	
\end{enumerate}
 \caption{\textsc{learnHKZ} \cite{Hsu2012}}
\end{algorithm}

\hspace{10 mm}

The underlying basic algorithm \textsc{learnHMM} by \cite{Hsu2012} has polynomial sample and computational complexity. Furthermore, assuming Conditions \ref{condition1HKZ} and \ref{condition2HKZ}, the basic algorithm yields a provably correct result. However, the algorithm does not ensure that the predicted probabilities lie in the range $\left[0,1\right]$ \cite{Hsu2012}.

\section{Spectral algorithm by Anandkumar et al.}
\label{sec:AHK}
This section will establish a learning algorithm as proposed in \cite{AHK2012}, including important mathematical derivations.\\
A simple bag-of-words model will be used to introduce cross moments and their representation using a tensor notation. This will then lead to the formulation of multi-view models and their application for hidden Markov models. Since the work by \cite{AHK2012} does not explicitly state a learning algorithm for HMMs, part \ref{subsec:derivations} will show important mathematical derivations. These derivations then lead to the spectral learning algorithm presented in part \ref{subsec:learnAHK}.\\
The presentation closely follows \cite{AHK2012} and \cite{AGHKT2012}. The proofs can be found in Appendix \ref{cha:appendix}.

\subsection{Bag-of-words}
\label{subsec:bagofwords}
The \textit{bag-of-words model} is a simple model where a document is assumed to be composed of a set of words in exchangeable order. Word ordering and grammar are therefore disregarded, but multiplicity of words is observed. Intuitively, in this model a document can be imagined as a bag full of unordered words.\\
The generative process can be described in two steps. First, the topic for a document is chosen as specified by a multinomial distribution out of the discrete set of distinct topics. Then, for the given topic, the document's words are drawn independently from the set of words, governed by another multinomial distribution.\\
Let there be a body of $k$ distinct topics a document can have. Let $d$ be the number of distinct words in the vocabulary and let $l$ be the number of words in a document. We assume $l \geq 3$.\\
The latent variable $h$ is identified with the topic of a given document. The probability for a document to have the topic $j \in \left[k\right]$ is given by
\[
	\Pr \left[ h=j\right] = w_j, \quad j \in \left[k\right],
\]
where $\vec{w} := \left( w_1,w_2,...,w_k \right) \in \Delta^{k-1}$ is a probability vector governing the distribution of topics. \\
For a given topic $h$, the probability vector $\vec{\mu}_h \in \Delta^{k-1}$ specifies a multinomial distribution from which the $l$ words of the documents are drawn. A practical way to represent the words is by using $d$-dimensional vectors. \\
Let the set of $l$ words in a document be $ \vec{x}_1,\vec{x}_2,...,\vec{x}_l \in \RR^d$. We then set
\[
	\vec{x}_t = \vec{e}_i \Leftrightarrow \text{the } t \text{-th word in the document is }i, \quad t \in \left[l\right], 
\]
where $\vec{e}_1,\vec{e}_2,...,\vec{e}_d$ is the standard coordinate basis for $\RR^d$. \\
This vector representation can now be used to intuitively describe joint probabilities over words.\\
 Let $\EE \left[ \vec{x}_1 \otimes \vec{x}_2 \right]$ be the matrix of joint probabilities over two words. The matrix is then given by
\begin{align*}
	\EE \left[ \vec{x}_1 \otimes \vec{x}_2 \right] &= 
				\sum_{1 \leq i,j \leq d} \Pr \left[ \vec{x}_1 = \vec{e}_i, \vec{x}_2 = \vec{e}_j\right] \vec{e}_i \otimes \vec{e}_j \\
		&=  \sum_{1 \leq i,j \leq d} \Pr \left[ 1^\text{st} \text{ word = } i, 2^\text{nd} \text{ word = }  j \right] \vec{e}_i \otimes \vec{e}_j .
\end{align*}
Therefore the $(i,j)$-th entry of the matrix $\EE \left[ \vec{x}_1 \otimes \vec{x}_2 \right]$ of joint probabilities is $\Pr \left[ 1^\text{st} \text{ word = } i, 2^\text{nd} \text{ word = }  j \right]$. This leads to a general tensor formulation for the joint probabilities over $l$ words:
\begin{multline*}
	\EE \left[ \vec{x}_1 \otimes \vec{x}_2 \otimes \cdots  \otimes \vec{x}_l \right] = \\
		\sum_{1 \leq i_1,i_2,\ldots,i_l \leq d} \Pr \left[ 1^\text{st} \text{ word = } i_1, 2^\text{nd} \text{ word = }  i_2,\ldots, l^\text{th} \text{ word = } i_l \right] \vec{e}_{i_1} \otimes \vec{e}_{i_2} \otimes \cdots \otimes \vec{e}_{i_l}.
\end{multline*}
The $(i_1,i_2,...,i_l)$-th entry of the tensor $\EE \left[ \vec{x}_1 \otimes \vec{x}_2 \otimes  \cdots  \otimes \vec{x}_l \right]$ corresponds to $\Pr \left[ 1^\text{st} \text{ word = } i_1, 2^\text{nd} \text{ word = }  i_2, ..., l^\text{th} \text{ word = } i_l \right]$. \\ 
An important result of this consideration is that the estimation of cross moments can be conducted by the estimation of joint probabilities of a sequence of words. As an example, the cross moment for $\vec{x}_1 \otimes \vec{x}_2 \otimes \vec{x}_3$ can be estimated by sampling the joint probabilities of the first three words over all documents. \\
Let us now consider the conditional expectation of $\vec{x}_t$ given a topic $h=j, \, j \in \left[ k \right]$:
\begin{align*}
	\EE \left[ \vec{x}_t | h=j \right] &= \sum_{i=1}^d \Pr \left[ t \text{-th word = } i | h=j \right] \vec{e}_i \\
	&= \sum_{i=1}^d \left[ \vec{\mu}_{j} \right]_i \vec{e}_i \\
	&= \vec{\mu}_{j}.
\end{align*}
The conditional expectation of a word $\vec{x}_t$ given a document topic $h=j$ is therefore equal the probability vector $\vec{\mu}_j \in \Delta^{k-1}, \, j \in \left[k\right]$. This is another important property that will be used in later considerations. \\

\subsection{Multi-view mixture models}
\label{subsec:multiview}
Coming from the bag-of-words model introduced in the previous section \ref{subsec:bagofwords}, a generalization now leads to the concept of \textit{multi-view models}. Similarly to the bag-of-words model, our observed variables $\vec{x}_1,\vec{x}_2,...,\vec{x}_l$ are assumed to be conditionally independent given a latent variable $h$. However, now the conditional distributions of the observations $\vec{x}_t, \, t \in \left[l\right]$ are not required to be identical anymore.\\
In a bag-of-words model, the topic of a document does not change while the words are picked according to their multinomial distribution. When considering the observations of a hidden Markov model however, the conditional distribution of an observation depends on the underlying hidden state. Since a HMM passes through a hidden state sequence, the state of the system might have changed from one observation to the subsequent one. \\
\subsubsection{General setting}
We now introduce the general setting for a multi-view mixture model. This setting needs to be established in order to properly introduce a learning algorithm as proposed by \cite{AHK2012}. In terms of its application to HMMs, the concept might not seem to be very intuitive at first. In the course of the formulation of the learning algorithm for HMMs in \ref{subsec:derivations}, the relationship between HMMs and the general quantities introduced here should become clear. To allow for a better understanding of the application of the multi-view mixture model to HMMs, the presentation here will be slightly different to \cite{AHK2012}. \\
We restrict our discussion to $x_t$ with $t \in \left\{ 1,2,3 \right\}$. This results in an instance of a three-view mixture model which will be the starting point for our considerations regarding hidden Markov models.

\begin{definition}
\label{def_general_setting} 
Let $k$ be the number of mixture components. \footnote{In the case of HMMs, this will later correspond to the number of states.} Let $l$ be the number of views, assuming $l \geq 3$. Let $\vec{w} = \left( w_1,w_2,...,w_k \right) \in \Delta^{k-1}$ be a vector of mixing weights, and let $h \in \left[k\right]$ be a hidden discrete random variable with $\Pr \left[ h=j \right] = w_j, \: \forall j \in \left[ k \right]$. Let $\vec{x}_1,\vec{x}_2,\vec{x}_3 \in \RR^d$ be random vectors that are conditionally independent given $h$. We define the conditional means of these vectors as 
\[
	\vec{\mu}_{t,j} := \EE \left[ \vec{x}_t | h=j \right], \quad j \in \left[ k \right], t \in \left\{ 1,2,3 \right\}.
\]
and $M_t \in \RR^{d \times k}, t \in \left\{ 1,2,3 \right\}$ as the matrix of conditional means, with the $j$-th column being $\vec{\mu}_{t,j}$:
\[
	M_t := \left[ \vec{\mu}_{t,1} | \vec{\mu}_{t,2} | \ldots | \vec{\mu}_{t,k} \right].
\]
\end{definition}


\subsection{Derivations}
\label{subsec:derivations}
The algorithm proposed in \cite{AHK2012} is formulated as a general method for learning multi-view mixture models. This section elaborates important considerations for the formulation as a spectral algorithm for HMMs. The resulting learning algorithm will then be presented in \ref{subsec:learnAHK}.\\
Using the setting introduced in Definition \ref{def_general_setting} we assume the following condition. 

\begin{condition}
\label{condition1}
	Non-degeneracy.\\ $w_j > 0$ for all $j \in \left[k\right]$, and $\operatorname{rank} \left(M_t\right)= k$ for all $t \in \left\{ 1,2,3 \right\}$.
\end{condition}

To make the transition from the general setting to hidden Markov models, we now relate the quantities $M_t, t \in \left\{ 1,2,3 \right\}$ to the parameters of a hidden Markov model.
\begin{proposition}
\label{prop_hmm}
\cite{AHK2012} If the hidden variable $h$ from the multi-view mixture model (Section \ref{subsec:multiview}) is identified with the second hidden state $h_2$, then $\left\{ \vec{x}_1, \vec{x}_2, \vec{x}_3 \right\} $ are conditionally independent given $h$, and the parameters of the resulting three-view mixture model on $\left(h, \vec{x}_1, \vec{x}_2, \vec{x}_3 \right)$ are
	\[
		\vec{w} = T\vec{\pi}, \quad M_1 = O \, \mathrm{diag} (\vec{w}) \, T^\top \, \mathrm{diag} (T\vec{\pi})^{-1}, 
			\quad M_2= O, \quad M_3 = OT.
	\]
\end{proposition}
For the following derivations we will however stick to the general notation using $M_1,M_2$ and $M_3$ at first, as the derivations are also valid for multi-view models in general. Furthermore this simplifies and shortens notation.

\begin{definition}
\label{defofmoments}
Second- and third-order moments.\\
We define the matrix of second-order moments $P_{3,1} \in \RR^{d \times d}$ and the tensor of third-order moments $P_{3,1,2} \in \RR^{d \times d \times d}$ by
	\begin{equation}
	\label{eq:momentdefinition}
		P_{3,1} := \EE \left[ \vec{x}_3 \otimes \vec{x}_1 \right] \quad \text{and}
			\quad P_{3,1,2} := \EE \left[ \vec{x}_3 \otimes \vec{x}_1 \otimes \vec{x}_2 \right].
	\end{equation}
\end{definition}
The tensor $P_{3,1,2}$ can also be regarded as a linear operator $P_{3,1,2}: \RR^d \rightarrow \RR^{d \times d}$ given by
\begin{equation}
\label{eq:linopdef}
	P_{3,1,2} (\vec{\eta}) := \EE \left[ \left( \vec{x}_3 \otimes \vec{x}_1 \right) \left\langle \vec{\eta}, \vec{x}_2 \right\rangle \right],
\end{equation}
where $ \left\langle \vec{x}, \vec{y}  \right\rangle = \vec{x}^\top \vec{y}$ denotes the standard inner product between two vectors. \\
For a vector $\vec{\eta} = \left( \eta_1,...,\eta_d \right)$ the $(i,j)$-th entry of the matrix $P_{3,1,2} (\vec{\eta})$ is then given by
\begin{equation}
\label{eq:linopvals}
	P_{3,1,2} (\vec{\eta})_{i,j} = \sum_{\alpha=1}^d \eta_\alpha \left[ P_{3,1,2} \right]_{i,j,\alpha},
\end{equation}
where $\left[ P_{3,1,2} \right]_{i,j,\alpha}$ is the $(i,j,\alpha)$-th element of the third-order moment tensor. \\

The second- and third-order moments in Definition \ref{defofmoments} can be related to the model parameters for multi-view mixture models. These relations will play an important role in formulating the parameter estimation algorithm. \\
The following lemma relates the moments defined above to the model parameters $M_t, t \in \left\{1,2,3\right\}$.
\begin{lemma}
\label{lem:secondthirdmoments}
	Let $P_{3,1}$ and $P_{3,1,2}$ be defined as in Definition \ref{defofmoments}. Then
	\[
		P_{3,1} = M_3 \, \mathrm{diag}(\vec{w}) \, M_1^\top \quad \mathrm{and}
			\quad P_{3,1,2} = M_3 \, \mathrm{diag}(M_2^\top \vec{\eta}) \, \mathrm{diag}(\vec{w}) \, M_1^\top \quad \forall \vec{\eta} \in \RR^d.
	\]
\end{lemma}
Given a sequence of observations, it is possible to efficiently sample the pair-wise and triple-wise probabilities to form empirical estimates for the second- and third-order moments. Lemma \ref{lem:secondthirdmoments}  now implies that these moments can be used to reveal the underlying model parameters. This is done using a so-called \textit{observable operator}. Observable operators were introduced by \cite{Jaeger2000} and are also used in the HMM learning algorithm by \cite{Hsu2012}.\\
The following important lemma introduces such an observable operator, a quantity that is only a function of joint probabilities of observable variables. Furthermore, the lemma relates this operator to our model parameters.
\begin{lemma}
\label{lem:observableoperator}
	\cite{AHK2012} Assume Condition \ref{condition1}. For $t \in \left\{ 1,2,3 \right\}$, let $U_t$ be a matrix such that $U_t^\top M_t$ is invertible. Then $U_3^\top P_{3,1} U_1$ is invertible, and for all $\vec{\eta} \in \RR^d$, the observable operator $B_{3,1,2} \in \RR^{k \times k}$, given by
	\begin{equation}
		B_{3,1,2}(\vec{\eta}) := (U_3^\top P_{3,1,2}(\vec{\eta}) U_1)(U_3^\top P_{3,1} U_1)^{-1}
	\end{equation}
	satisfies
	\begin{equation}
	\label{eq:empiricaloperator}
		B_{3,1,2} (\vec{\eta})  = \left( U_3^\top M_3 \right) \mathrm{diag} \left( M_2^\top \vec{\eta} \right) \left( U_3^\top M_3 \right)^{-1}.
	\end{equation}
In particular, the $k$ roots of the polynomial $\lambda \mapsto \mathrm{det} \left( B_{3,1,2}(\vec{\eta}) - \lambda I \right)$ are $\left\{ \left\langle \vec{\eta},\vec{\mu}_{2,j} \right\rangle : j \in \left[ k \right] \right\}$.
\end{lemma}
The matrix $B_{3,1,2} (\vec{\eta})$ is therefore similar to the diagonal matrix $\mathrm{diag} \left( M_2^\top \vec{\eta} \right)$, and the columns of $U_3^\top M_3$ are the eigenvectors of $B_{3,1,2} (\vec{\eta})$. Considering the case where $\vec{\eta} = \vec{e}_i, i \in \left[ d \right]$, this would now relate our desired model parameters to our observable operator matrix. In absence of prior knowledge, these vectors $\vec{\eta} = \vec{e}_i$ are not known though. It is, however, possible to obtain the model parameters by applying different observable operators to different vectors $\vec{\eta}$:\\

\begin{lemma}
\label{lem:paramrecovery}
 \cite{AHK2012} Consider the setting and definitions from Lemma \ref{lem:observableoperator}. Let $\Theta \in \RR^{k \times k}$ be an invertible matrix, and let $\vec{\theta}_i \in \RR^k$ be its $i$-th row. For all $i \in \left[ k\right]$, let $\lambda_{i,1},\lambda_{i,2},...,\lambda_{i,k}$ denote the $k$ eigenvalues of $B_{3,1,2} ( U_2 \vec{\theta}_i )$ in the order specified by the matrix of right eigenvectors $U_3^\top M_3$. Let $L \in \RR^{k \times k}$ be the matrix whose $(i,j)$-th entry is $\lambda_{i,j}$. Then
\begin{equation}
\label{eq:parametersystem}
		\Theta U_2^\top M_2 = L.
\end{equation}
\end{lemma}
This lemma now shows that the unknown parameters $M_2$ can be obtained by the solution of the linear system \eqref{eq:parametersystem}. The elements of the matrix $L$ are the roots of the $k$-th degree polynomial of the observable operator matrices derived from the empirically observed second- and third-order moments. Proposition \ref{prop_hmm} now states that $M_2 = O$, so we now have established a method to learn the emission probability matrix:
\begin{equation}
	O = U_2 \Theta^{-1} L.
\end{equation}
The obtained matrix $O$ together with Lemma \ref{lem:observableoperator} and Proposition \ref{prop_hmm} now enables us to calculate an estimate for the state transition probability matrix $T$. Using Equation \eqref{eq:empiricaloperator} and inserting the quantities $M_3$ and $M_2$ from Proposition \ref{prop_hmm} we obtain
\begin{align*}
	B_{3,1,2} (\vec{\eta})  &= \left( U_3^\top M_3 \right) \mathrm{diag} \left( M_2^\top \vec{\eta} \right) \left( U_3^\top M_3 \right)^{-1} \\
	& = \underbrace{\left( U_3^\top O T \right)}_{=: R} \mathrm{diag} \left( O^\top \vec{\eta} \right) \left(U_3^\top O T \right)^{-1} \\
	& = R \, \mathrm{diag} \left( O^\top \vec{\eta} \right)  R^{-1},
\end{align*}
where $R$ is the matrix of right eigenvectors of $B_{3,1,2}(\vec{\eta})$. Therefore, our state transition probability matrix is given by
\begin{equation}
	T = \left( U_3^\top O \right)^{-1} R.
\end{equation}

\subsection{Algorithm learnAHK}
\label{subsec:learnAHK}
The spectral learning algorithm is now the result of the relations revealed by the derivations in \ref{subsec:derivations}. The lemmas suggest the following estimation procedure:\\
Proposition \ref{prop_hmm} shows how the parameters of a hidden Markov model relate to the moments in a multi-view (more specifically, three-view) mixture model. Lemma \ref{lem:secondthirdmoments} relates these moments to quantities that can be obtained empirically, given a sequence of observations. The observable operators introduced in Lemma \ref{lem:observableoperator} can then be used to reveal our model parameters by a simultaneous diagonalization as implied by Lemma \ref{lem:paramrecovery}.\\
Note that due to the nature of the singular value decomposition, the ordering of the columns of the estimated quantities is subject to permutation. The estimated transition probability matrix is only calculated up to a scaling of the columns. The implementation of the algorithm in this work performed a basic normalization to obtain transition probabilities that sum to $1$.\\
The framework of the algorithm is suitable for both categorical observations and continuous multivariate observations.

\newpage
\begin{algorithm}[H]
\label{alg:learnAHK}
\SetAlgoLined
 \KwData{$N$ triples of observations, $k$ - number of states , $d$ - number of observations}
 \KwResult{Hidden Markov model parameterized by $\hat{O}$ and $\hat{T}$}

\begin{enumerate}
	\item Obtain empirical estimates for $\hat{P}_{3,1} \in \RR^{k \times k}$, $\hat{P}_{3,2} \in \RR^{k \times k}$ and $\hat{P}_{3,1,2} \in \RR^{k \times k \times k}$ from the $N$ observation triples.
	\item For the largest $k$ singular values of $\hat{P}_{3,1}$, find the matrix $\hat{U}_3 \in \RR^{d \times k}$ of orthonormal left and the matrix $\hat{U}_1 \in \RR^{d \times k}$ of orthonormal right singular vectors. \\
			For the largest $k$ singular values of $\hat{P}_{3,2}$, find the matrix $\hat{U}_2 \in \RR^{d \times k}$ of orthonormal right singular vectors.
	\item Pick an invertible matrix $\Theta \in \RR^{k \times k}$, with the $i$-th row denoted as $\vec{\theta}_i^\top \in \RR^k$. In the absence of prior knowledge about $\hat{O}$, choose $\Theta$ to be a random rotation matrix. \\
			For all $i \in \left[ k \right]$ calculate the matrix $\hat{P}_{3,1,2} (\hat{U}_3 \vec{\theta}_i )$, with entries
				\[
					\hat{P}_{3,1,2} (\hat{U}_3 \vec{\theta}_i )_{k,l} = \sum_{\alpha=1}^d \left[ \hat{U}_3 \vec{\theta}_i \right]_\alpha \left[ P_{3,1,2} \right]_{k,l,\alpha}.
				\]
			For all $i \in \left[ k \right]$, form the matrices
				\[
					\hat{B}_{3,1,2}(\hat{U}_3 \vec{\theta}_i) = (\hat{U}_3^\top \hat{P}_{3,1,2}(\hat{U}_3 \vec{\theta}_i) \hat{U}_1)(\hat{U}_3^\top \hat{P}_{3,1} \hat{U}_1)^{-1}
				\]
			Diagonalize the matrix $\hat{B}_{3,1,2}(\hat{U}_3^\top \vec{\theta_1})$ such that
				\[
					\hat{R}_3^{-1} \hat{B}_{3,1,2}(\hat{U}_3^\top \vec{\theta_1}) \hat{R}_3 = \mathrm{diag} \left( \hat{\lambda}_{1,1}, \hat{\lambda}_{1,2},\ldots,\hat{\lambda}_{1,k} \right).
				\]
				(Fail if not possible.)
	\item For each $i \in \left\{ 2,\ldots,k \right\}$, obtain diagonal entries $\hat{\lambda}_{i,1}, \hat{\lambda}_{i,2},\ldots,\hat{\lambda}_{i,k}$ of $\hat{R}_3^{-1} \hat{B}_{3,1,2}(\hat{U}_3^\top \vec{\theta_i}) \hat{R}_3$ and form the matrix $\hat{L} \in \RR^{k \times k}$ with the $(i,j)$-th entry being $\hat{\lambda}_{i,j}$.
	\item Return $\hat{O} = \hat{U}_2 \Theta^{-1} \hat{L}$.
	\item Return $\hat{T} = \left( \hat{U}_3 \hat{O} \right)^{-1} \hat{R}_3$, scale columns of $\hat{T}$.
	
\end{enumerate}
 \caption{\textsc{learnAHK} \cite{AHK2012}}
\end{algorithm}


\section{Real-valued emissions}
\label{sec:real_valued_section}
When considering experiments in real-world applications, measurement data is not likely to be categorical. Furthermore, measurements will suffer from both intrinsic noise in the underlying processes and extrinsic noise induced by the measurement setup. \\
Let there be a time series of real-valued emissions, $\left\{y_t\right\}_{t = 1,2,\ldots,L}$ with $y_t \in \RR, \forall t$. Assume that the generative process of our measurement data can be described by a hidden Markov model. We now want to use the spectral algorithms (Section \ref{sec:HKZalg} and Section \ref{sec:AHK}) to learn parameter estimates from this data. The algorithms do not allow the input of the real-valued sequences directly. Thus, the challenge is to find a representation for our real-valued emissions so that they can be used as an input for the spectral algorithms.\\
One possible representation would be to obtain a sequence of categorical emissions by binning the real-valued emissions. The approach evaluated here will be to bin by quantiles. Taking all the emissions available, quantiles can be calculated, leading to natural bin bounds.\\
A simple binning approach is introduced in this section. The approach will then be evaluated in Section \ref{sec:real_valued_results}.
\subsection{Simple binning approach}
\label{subsec:simple_binning_theory}
Assume we are given a sequence of real-valued observations. Furthermore, assume that the dimension, or number, of emissions $d$ of the hidden Markov model is given. One of the simplest approaches to prepare the data for the learning algorithms is to bin the observations to generate a sequence of categorical emissions. The major challenge is to find appropriate bin bounds.\\
The approach pursued in this method is to use all available observations and calculate quantiles. These quantiles then yield natural bounds that are used for binning. The procedure is formulated in the following algorithm:\\

\begin{algorithm}[H]
\SetAlgoLined
 \KwData{Sequence of real-valued observations, $d$ - dimension of observations}
 \KwResult{Sequence of binned observations}
\begin{enumerate}
	\item Using the real-valued observations, calculate the quantiles for $d-1$ evenly spaced cumulative probabilities $\left( \frac{1}{d}, \frac{2}{d}, \ldots, \frac{d-1}{d} \right)$
	\item Observing the ordering of the sequence, bin each observation according to the bounds specified by the quantiles.
	\item Return sequence of binned observations.
\end{enumerate}
 \caption{\textsc{SimpleBinning}}
\end{algorithm}
\hspace{8 mm}

The sequence of categorical variables obtained by \textsc{SimpleBinning} can then be used as the input for the spectral learning algorithms \textsc{learnHKZ} and \textsc{learnAHK}. This procedure allows the learning of HMM model parameters given a sequence of real-valued emissions.\\
There are, however, intrinsic problems arising from this simple binning approach. Quantiles are calculated using all available observations. The resulting bin bounds will divide the total of all real-valued observations into bins of equal size. In our categorical sequence resulting from the binning, each possible observation will therefore occur equally often across all elements of the sequence.
When considering the HMM, this would correspond to the assumption that the cumulative probability of each of the possible emissions is the same. Without loss of generality, this assumption cannot be sustained.\\
The evaluation in Section \ref{sec:real_valued_results} will address these issues and provide ideas for improved representations of real-valued emissions for the use in spectral algorithms.

\chapter{Results and discussion}
\label{cha:results_and_discussion}
This chapter will present and discuss the experiments conducted in this work. Section \ref{sec:comparison_algorithms} characterizes the spectral learning algorithms. Using synthetic experimental data, the algorithms \textsc{learnHKZ} and \textsc{learnAHK} will be compared with each other. Furthermore, the algorithm \textsc{learnAHK} will be compared to the Baum-Welch algorithm. The Baum-Welch algorithm was chosen as an example of a common iterative method for learning hidden Markov models. Section \ref{sec:real_valued_results} then addresses how real-valued emissions can be used for learning parameter estimates using the spectral algorithms.

\section{Comparison of spectral learning algorithms}
\label{sec:comparison_algorithms}

\subsection{Learning synthetic HMMs}
The algorithms \textsc{learnHKZ} and \textsc{learnAHK} were evaluated using synthetic experimental data generated from four different example hidden Markov models. The first two models have a state dimension of $k=2$ and the second pair of models has a state dimension of $k=3$. For the latter two models, a larger number of possible emissions was chosen. The parameters of the models used are

\begin{tabular}{c | c}
$\mathbf{k=2}$, $\mathbf{d=3}$ & $\mathbf{k=2}$, $\mathbf{d=6}$ \\ \hline
$T =
 \begin{pmatrix}
  9/10 & 3/10 \\
  1/10 & 7/10
 \end{pmatrix} $ & 
$T =
 \begin{pmatrix}
  9/10 & 1/20 \\
  1/10 & 19/20
 \end{pmatrix} $
\\[0.4cm]
$ O =
	\begin{pmatrix}
  1/4 & 8/10 \\
  1/2 & 1/10 \\
	1/4 & 1/10 
 \end{pmatrix} $ & 
$ O =
	\begin{pmatrix}
  1/6 & 7/12 \\
  1/6 & 1/12 \\
	1/6 & 1/12 \\
	1/6 & 1/12 \\
	1/6 & 1/12 \\
	1/6 & 1/12
 \end{pmatrix} $
 \\[0.4cm]
$ \vec{\pi} = \left( 4/5, 1/5 \right) ^\top $ & 
$ \vec{\pi} = \left( 3/4, 1/4 \right) ^\top $
\end{tabular}\\[0.5cm]

and

\begin{tabular}{c | c}
$\mathbf{k=3}$, $\mathbf{d=8}$ & $\mathbf{k=3}$, $\mathbf{d=10}$ \\ \hline
$T =
 \begin{pmatrix}
  8/10 & 1/15 & 1/8 \\
  1/10 & 13/15 & 1/8 \\
	1/10 & 1/15 & 3/4 
 \end{pmatrix} $ & 
$T =
 \begin{pmatrix}
  8/10 & 1/15 & 1/6 \\
  1/10 & 13/15 & 1/6 \\
	1/10 & 1/15 & 2/3 
 \end{pmatrix} $
\\[0.7cm]
$ O =
	\begin{pmatrix}
  3/10 & 1/20 & 1/50 \\
	1/10 & 13/20 & 1/50 \\
	1/10 & 1/20 & 22/50 \\
	1/10 & 1/20 & 22/50 \\
	1/10 & 1/20 & 1/50 \\
	1/10 & 1/20 & 1/50 \\
	1/10 & 1/20 & 1/50 \\
	1/10 & 1/20 & 1/50
 \end{pmatrix} $ & 
$ O =
	\begin{pmatrix}
  6/15 & 1/20 & 1/50 \\
	1/15 & 11/20 & 1/50 \\
	1/15 & 1/20 & 21/50 \\
	1/15 & 1/20 & 21/50 \\
	1/15 & 1/20 & 1/50 \\
	1/15 & 1/20 & 1/50 \\
	1/15 & 1/20 & 1/50 \\
	1/15 & 1/20 & 1/50 \\
	1/15 & 1/20 & 1/50 \\
	1/15 & 1/20 & 1/50
 \end{pmatrix} $ 
 \\[0.4cm]
$ \vec{\pi} = \left( 1/3, 1/3, 1/3 \right) ^\top $ & 
$ \vec{\pi} = \left( 1/3, 1/3, 1/3 \right) ^\top $
\end{tabular}\\[0.5cm]
The model for $k=2$, $d=6$ was taken from the MathWorks Matlab R2012b documentation. The matrices of all four models were chosen due to their low condition numbers. \footnote{The condition number of a matrix is the ratio of the largest singular value to the smallest singular value  of that matrix.}\\
From each model, $N$ sequences of triples of categorical emissions were generated, with 
\[N = 1000,2500,5000,10000,25000,50000,100000. \] 
These sequences of triples were then used as the input for the learning algorithms. For each $N$, 100 realizations of the sequences were generated. \\
Both algorithms return empirical estimates for the state transition probability matrix $\hat{T}$ and the emission probability matrix $\hat{O}$. These empirical estimates were compared to the known true values for $T$ and $O$ under the squared Frobenius norm $\| \cdot \|_F^2$. Furthermore, the running times of the algorithms were recorded. The 100 realizations then allowed to calculate average estimation errors and average running times for the different learning instances.\\
All simulations were implemented using the software Matlab on a personal computer platform. The specifications of the computing environment can be found in Appendix \ref{sec:appendix_specs}. The source code for the implementations of the algorithms \textsc{learnHKZ} and \textsc{learnAHK} is available online, see Appendix \ref{sec:source_code}.\\

\subsection{Accuracy and dimensionality}
\label{sec:accuracy_and_dimensionality}
The two spectral algorithms were compared with respect to the accuracy of the estimated model parameters and their running time. Figure \ref{fig:comparek2} shows the results for two-state HMMs, and Figure \ref{fig:comparek3} shows the results for three-state HMMs.\\

\subsubsection{Two-state HMMs}
\begin{figure}
        \centering
        \begin{subfigure}[b]{0.5\textwidth}
                \centering\includegraphics[scale=0.7]{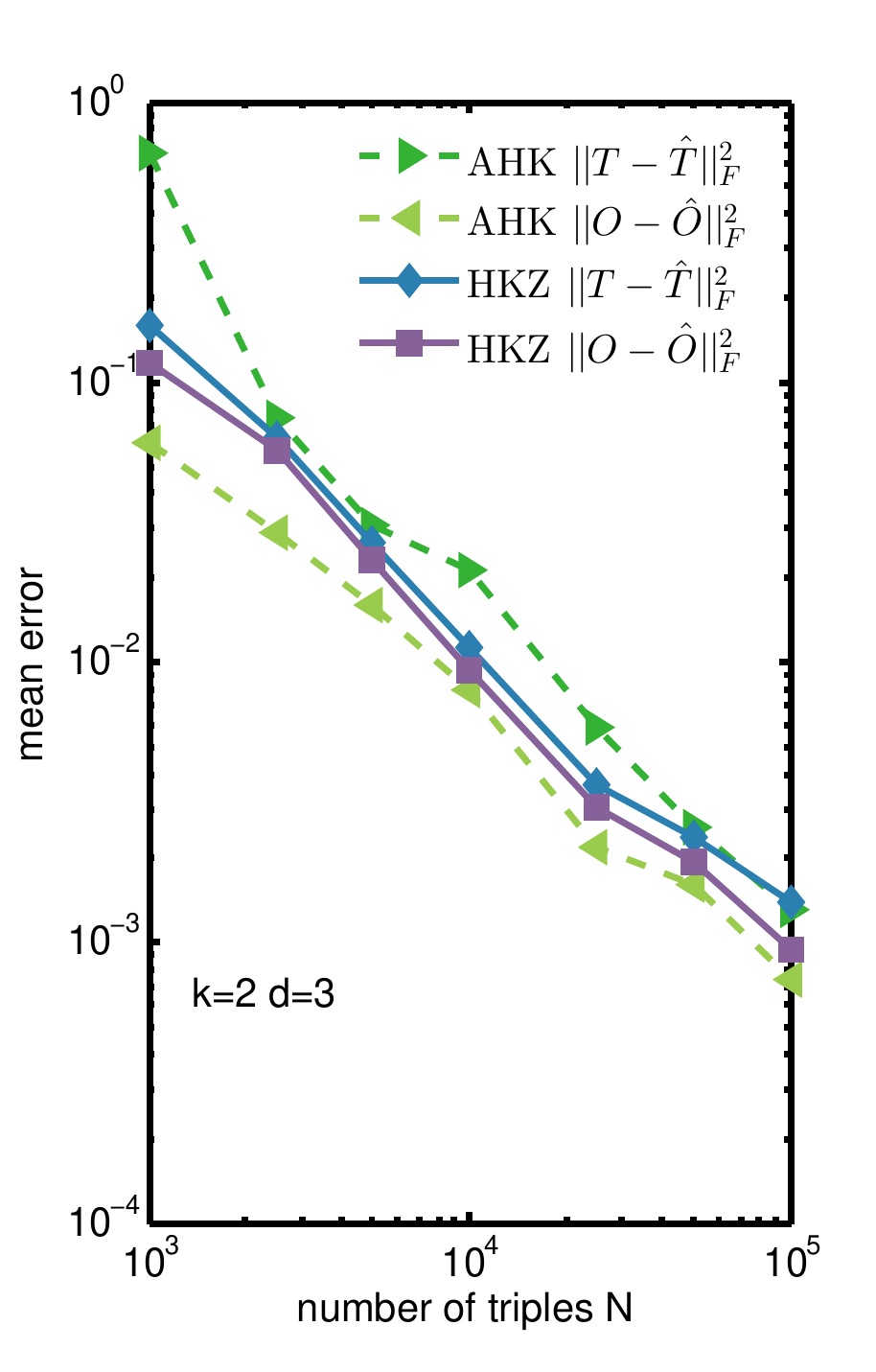}
                \caption{Mean errors $\mathbf{k=2}$, $\mathbf{d=3}$.}
								\label{fig:comparek2a}
        \end{subfigure}%
        \begin{subfigure}[b]{0.5\textwidth}
                \centering\includegraphics[scale=0.7]{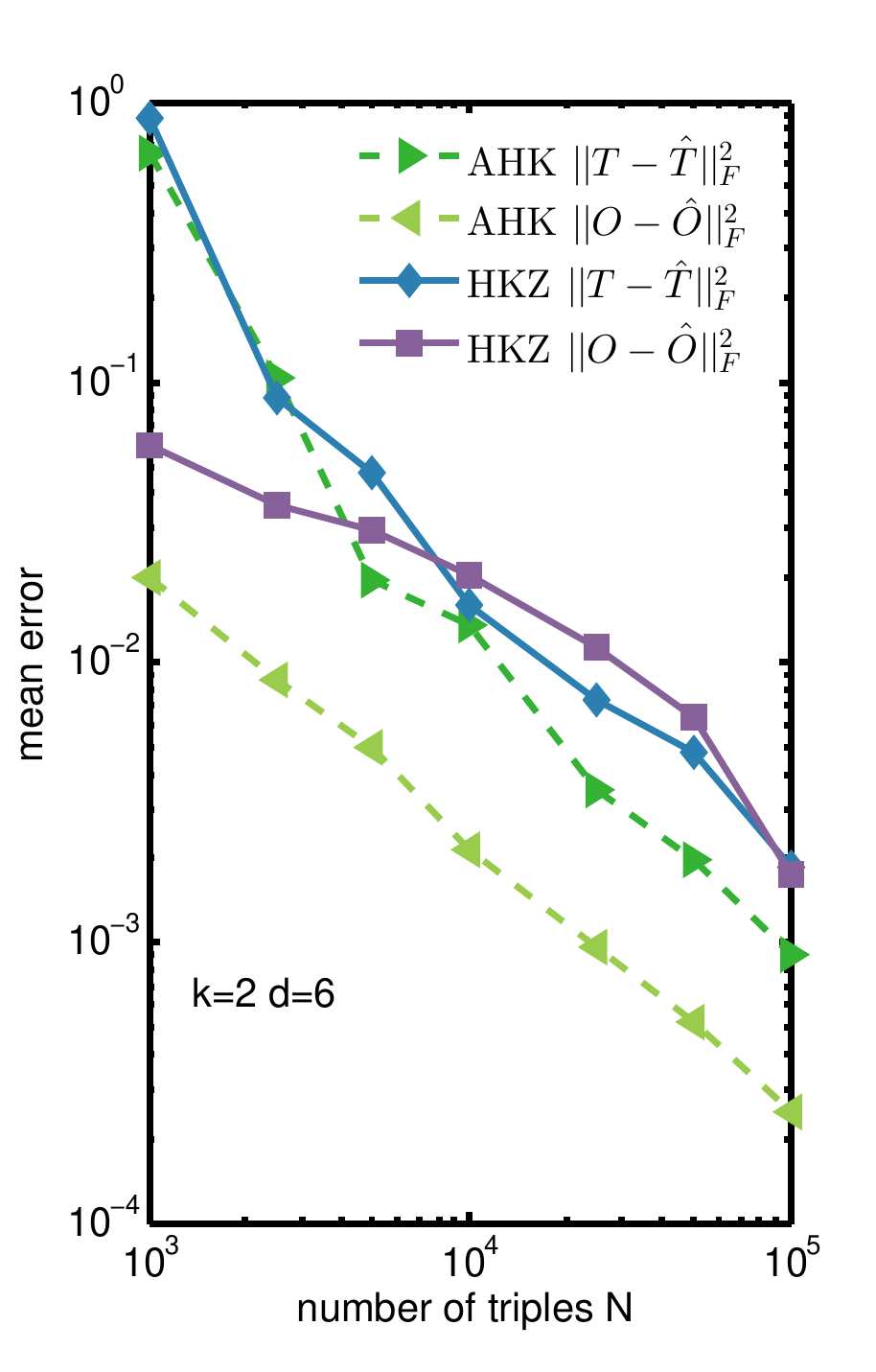}
                \caption{Mean errors $\mathbf{k=2}$, $\mathbf{d=6}$.}
                \label{fig:comparek2b}
        \end{subfigure} \\
				\begin{subfigure}[b]{\textwidth}
                \centering\includegraphics[scale=0.75]{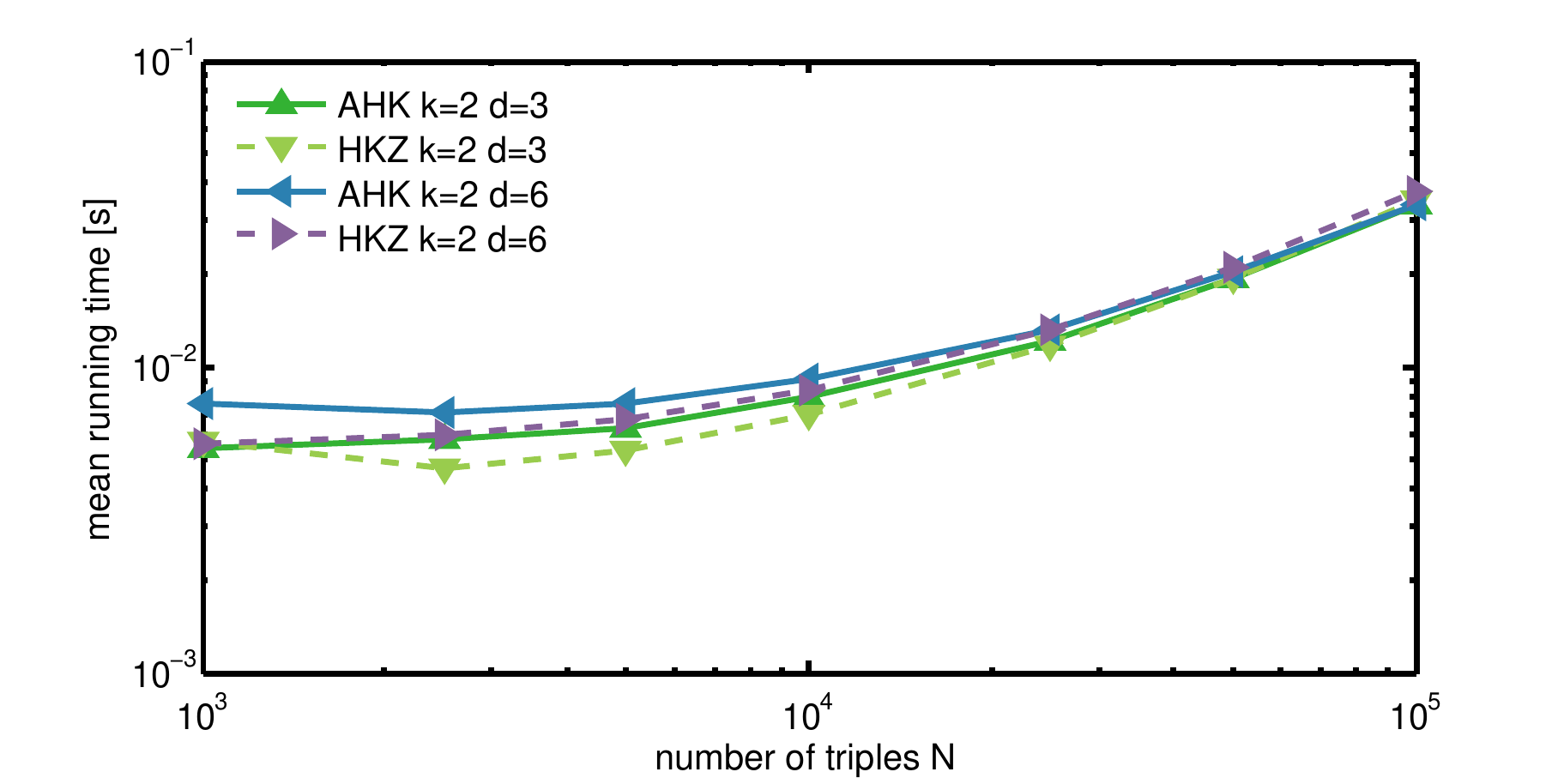}
                \caption{Mean running time for $\mathbf{k=2}$.}
                \label{fig:comparek2c}
        \end{subfigure}
        \caption{Comparison of the accuracy and running time of \textsc{learnHKZ} and \textsc{learnAHK} for $\mathbf{k=2}$.}
				\label{fig:comparek2}
\end{figure}
For the two-state hidden Markov model with three emissions (Figure \ref{fig:comparek2a}), both algorithms show the same level of accuracy. Furthermore the accuracy significantly increases with the number of triples of observations that used for learning.\\
When learning from experimental data sampled from the two-state HMM with six emissions, the algorithm \textsc{learnAHK} gives more accurate estimates than \textsc{learnHKZ}, see Figure \ref{fig:comparek2b}.\\
Figures \ref{fig:comparek2a} and \ref{fig:comparek2b} show the data on a $\mathrm{log}-\mathrm{log}$ scale. The average slopes of the estimation errors are given in the following table:\\

\begin{tabular}{ c | c c}
   & $k=2$, $d=3$ & $k=2$, $d=6$ \\ \hline
  \textsc{learnHKZ} & -1.08 & -0.91 \\
  \textsc{learnAHK} & -1.06 & -1.03 \\
\end{tabular}\\ \\
The slopes were obtained using a linear fit on the data over the sample size $N$, excluding $N=1000$.\\
The estimation error of both learning algorithms therefore decreases approximately exponentially with the sample size.\\
Figure \ref{fig:comparek2c} shows the running time of both algorithms. Also depicted on a $\mathrm{log}-\mathrm{log}$ scale, both algorithms show equal running time, increasing polynomially with the sample size.

\subsubsection{Three-state HMMs}
\begin{figure}
        \centering
        \begin{subfigure}[b]{0.5\textwidth}
                \centering\includegraphics[scale=0.7]{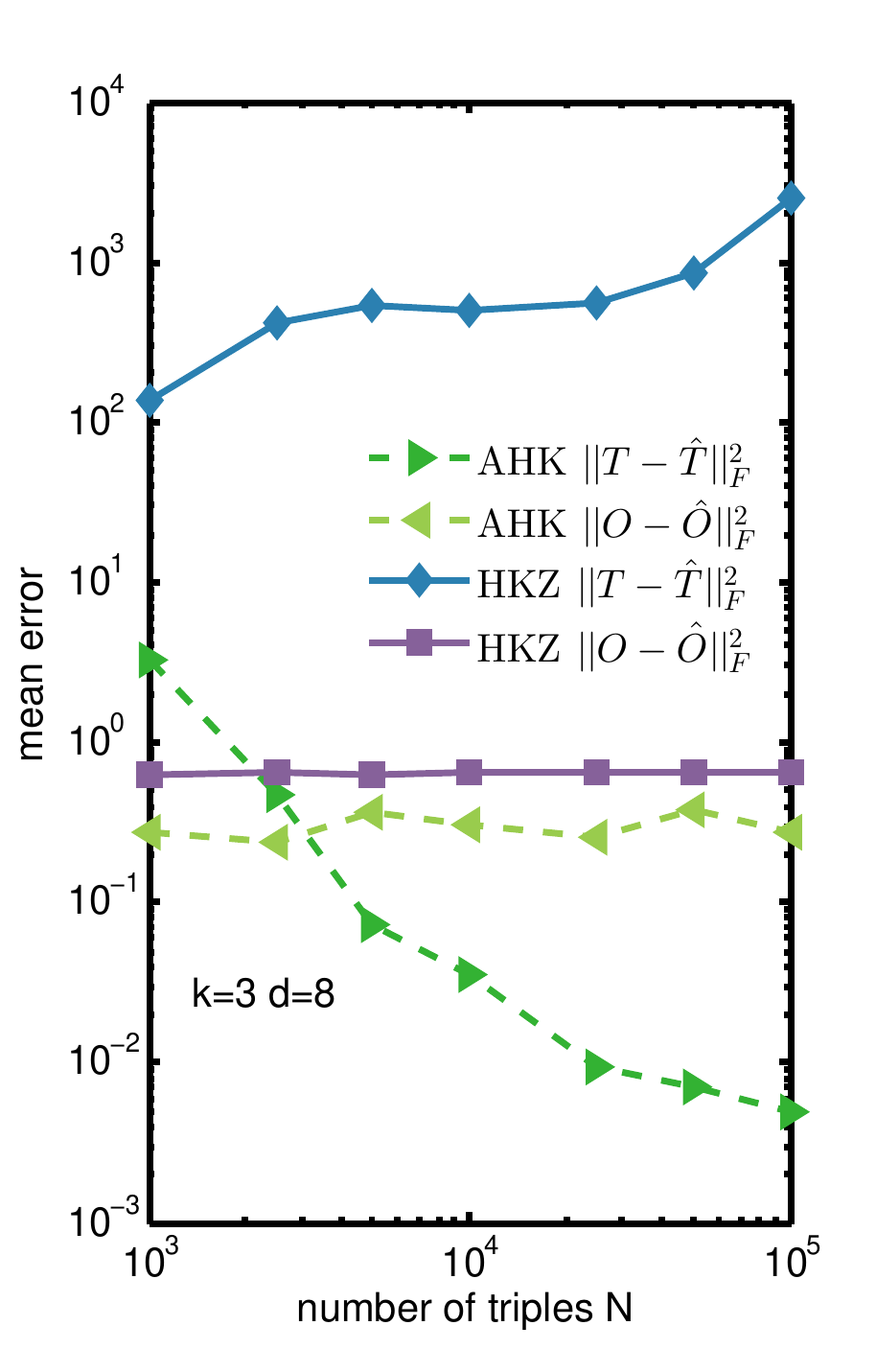}
                \caption{Mean errors $\mathbf{k=3}$, $\mathbf{d=8}$.}
								\label{fig:comparek3a}
        \end{subfigure}%
        \begin{subfigure}[b]{0.5\textwidth}
                \centering\includegraphics[scale=0.7]{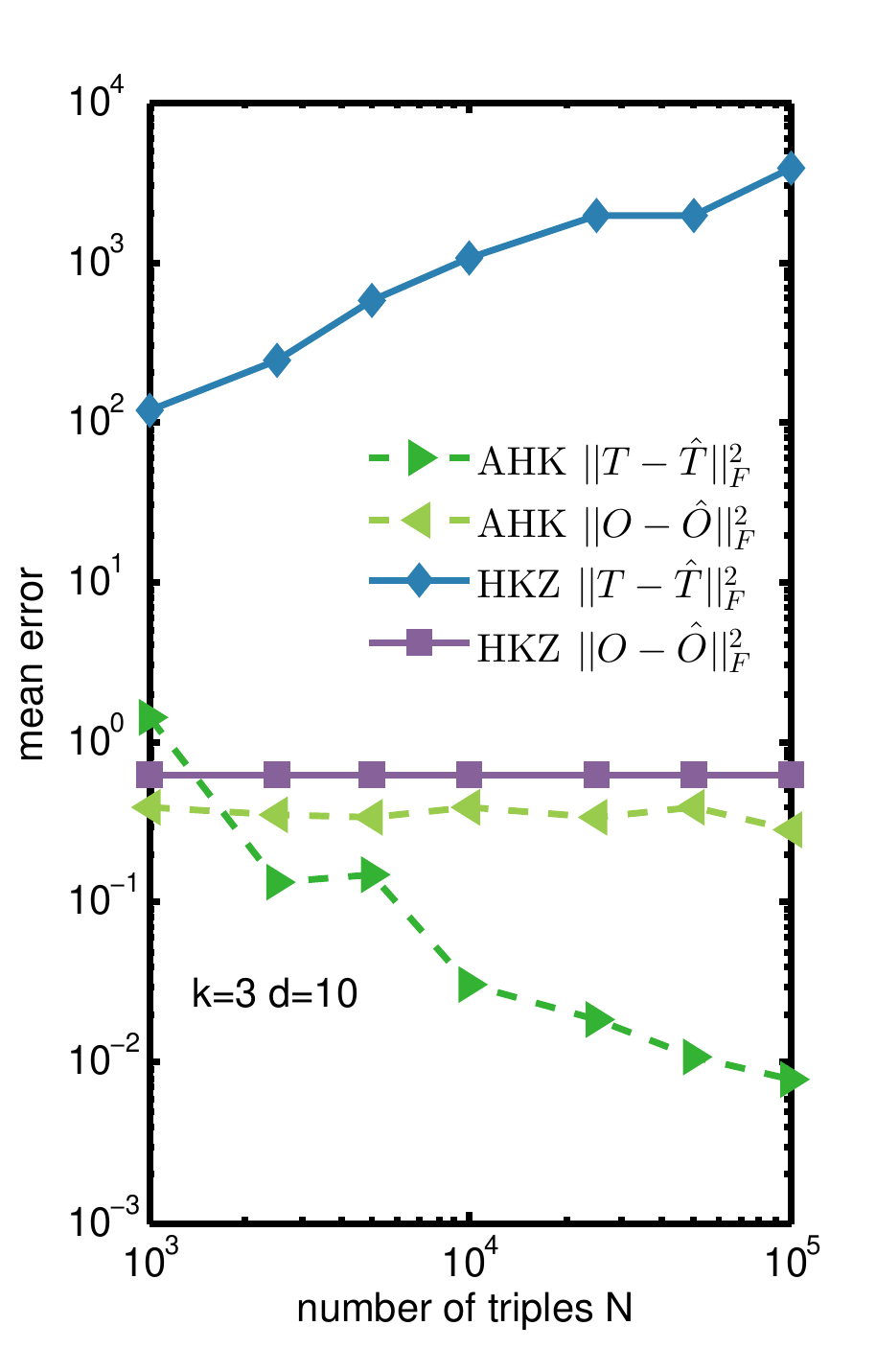}
                \caption{Mean errors $\mathbf{k=3}$, $\mathbf{d=10}$.}
                \label{fig:comparek3b}
        \end{subfigure} \\
				\begin{subfigure}[b]{\textwidth}
                \centering\includegraphics[scale=0.75]{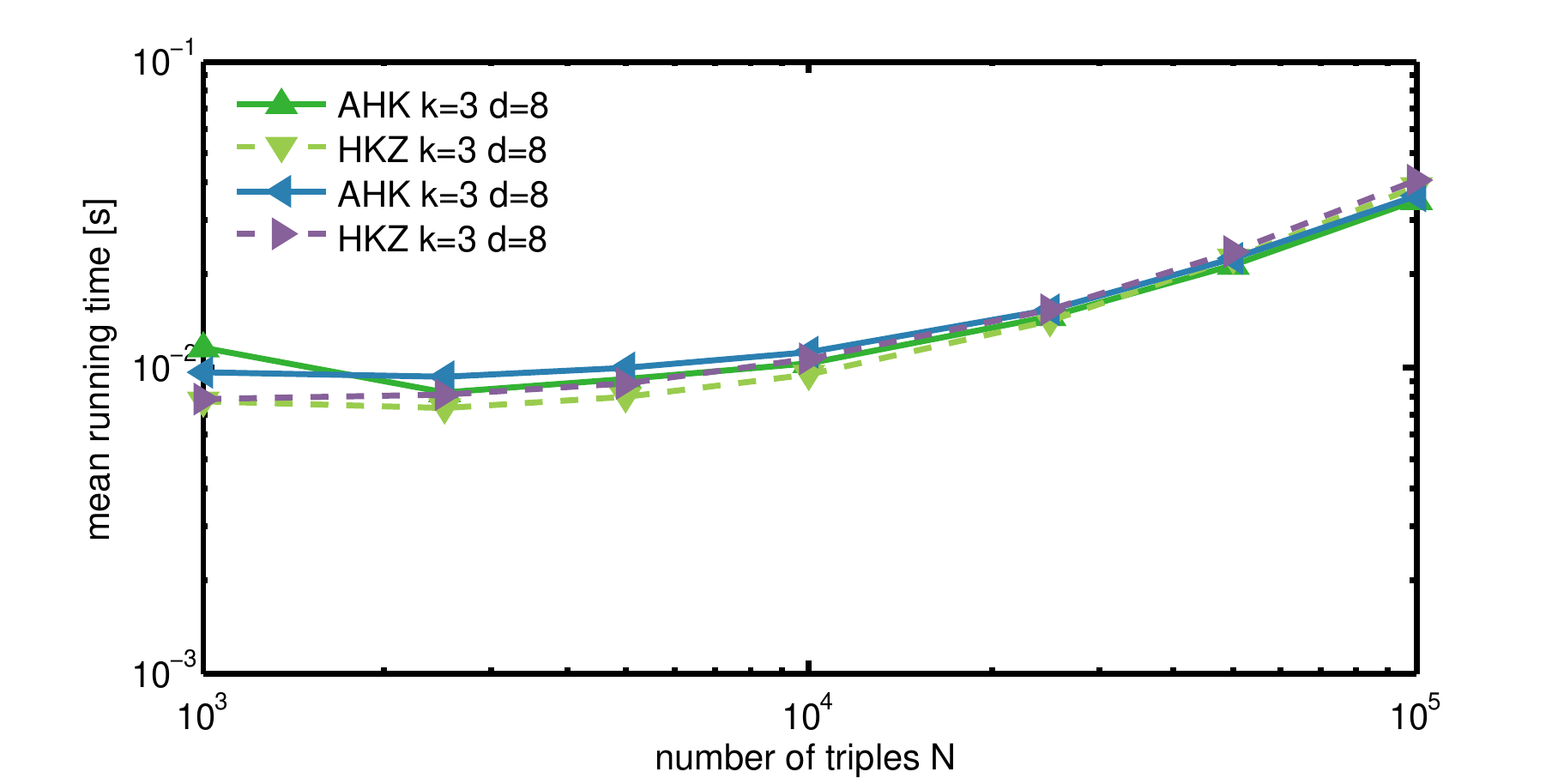}
                \caption{Mean running time for $\mathbf{k=3}$.}
                \label{fig:comparek3c}
        \end{subfigure}
        \caption{Comparison of the accuracy and running time of \textsc{learnHKZ} and \textsc{learnAHK} for $\mathbf{k=3}$.}
				\label{fig:comparek3}
\end{figure}
In a second step, the algorithms were compared using data from three-state HMMs with eight and ten possible emissions. The results are shown in Figure \ref{fig:comparek3}.\\
The estimation errors for learning the three-state eight-emission and three-state ten-emission HMM are depicted in Figures \ref{fig:comparek3a} and \ref{fig:comparek3b}, respectively. In contrast to the two-state HMMs with a lower number of possible emissions, the algorithm \textsc{learnHKZ} does not show a decreasing estimation error with increasing sample size. The estimation error for learning the transition probability matrix $\hat{T}$ even increases with the sample size. The algorithm \textsc{learnAHK} only shows an improving accuracy for $\hat{T}$. For both models and both algorithms, the estimation error of $\hat{O}$ does not decrease with increasing sample size.\\

This behavior of the learning algorithms is not expected. Ideally, the accuracy should improve for larger sample sizes, just as observed for the two-state HMMs above. The matrices used to generate the observation sequences were chosen to be well-conditioned. However, the matrix decompositions and inversions in the algorithms are prone to instability when the experimentally obtained matrices are ill-conditioned. Especially for matrices with high dimensionality, instability during the numerical operations used in these algorithms is an omnipresent issue. \\
The decreasing accuracy for the empirical estimate $\hat{T}$ observed for the algorithm \textsc{learnHKZ} could be due to another issue: the method used to obtain the parameter estimates does neither guarantee probabilities in the range $\left[ 0,1 \right]$, nor does it observe the scaling of the columns of the transition probability matrix that is estimated \cite{Hsu2012}. Both these issues are likely to be contributing to this unusual behavior.\\
The running time for both algorithms, depicted in Figure \ref{fig:comparek3c}, is approximately the same. This is consistent with the results for two-state HMMs.\\

\subsection{Comparison with the Baum-Welch algorithm}
\begin{figure}
        \centering
        \begin{subfigure}[b]{\textwidth}
                \centering\includegraphics[scale=0.65]{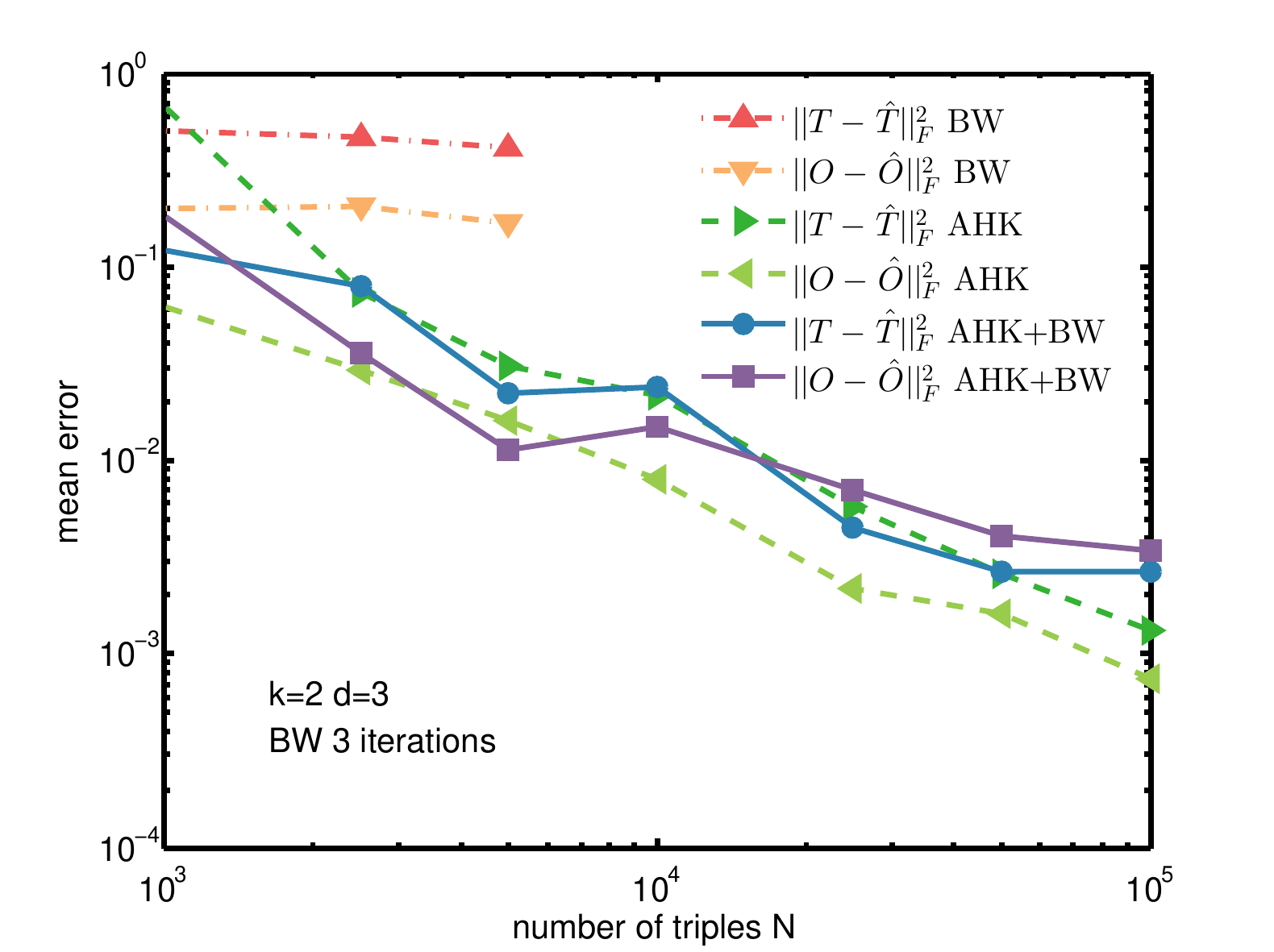}
                \caption{Mean errors for $\hat{O}$ and $\hat{T}$}
								\label{fig:compareBWa}
        \end{subfigure}%
				\\
        \begin{subfigure}[b]{\textwidth}
                \centering\includegraphics[scale=0.7]{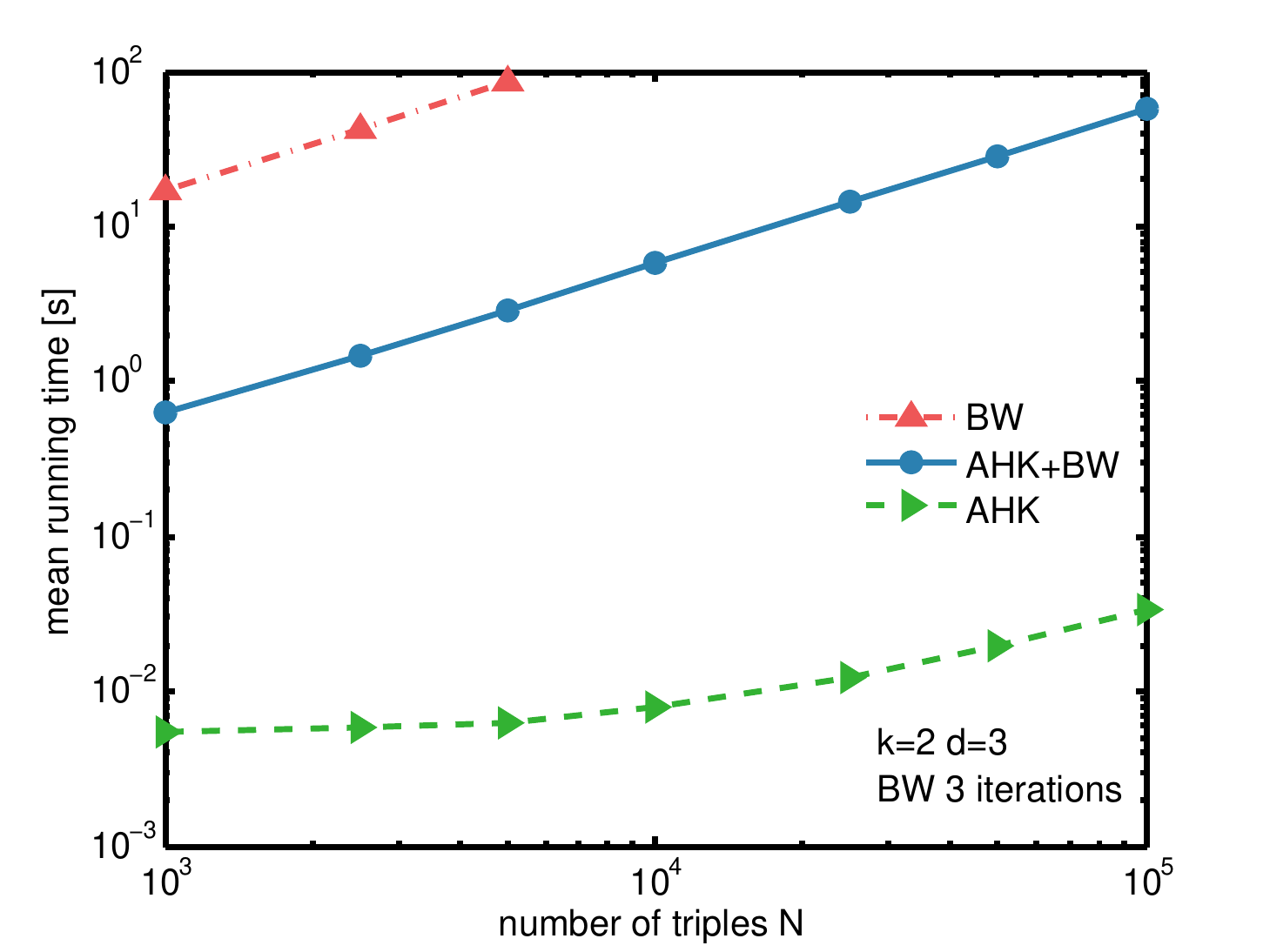}
                \caption{Mean running times.}
                \label{fig:compareBWb}
        \end{subfigure}
        \caption{Comparison of the algorithm \textsc{learnAHK} (indicated by \textit{AHK}) with the Baum-Welch algorithm, both with a random initialization (\textit{BW}), and using initial values from the \textsc{learnAHK} algorithm (\textit{AHK+BW}). Average errors and running times across 100 realizations for \textit{AHK} and 10 realizations for \textit{BW} and \textit{AHK+BW}}
				\label{fig:compareBW}
\end{figure}
In a next step, the algorithm \textsc{learnAHK} was compared to the Baum-Welch algorithm. For the simulations, the implementation of the Baum-Welch algorithm as provided by the Matlab Statistics Toolbox was used. In this experiment, only sequences sampled from the two-state three-emission ($k=2$, $d=3$) hidden Markov model specified earlier were considered. Using the same model experimental data as before, the Baum-Welch (BW) algorithm was run for learning the parameter estimates of the HMM.\\
The spectral algorithms do not require initial estimates for the quantities to be learned. With the BW algorithm being an iterative method, however, initial estimates are required for initialization. Therefore, two different scenarios were considered: the BW algorithm with a random initialization, and the BW algorithm initialized with the parameters learned by \textsc{learnAHK}.\\
The results are shown in Figure \ref{fig:compareBW}. For the BW algorithm with random initialization, only sample sizes $N = 1000,2500,5000$ were considered, due to the long running time. In both instances, the BW algorithm was run for three iterations. The results for the BW algorithm are averages over 10 realizations of input sequences. \\
Figure \ref{fig:compareBWb} shows the running time of the different learning instances. It can be seen that already for the smallest sample size, $N=1000$, the BW algorithm is much slower than the spectral algorithm. The difference in running time is more than three orders of magnitude. Compared to the BW algorithm initialized with learning results from \textsc{learnAHK}, the spectral algorithm by itself is still two orders of magnitude faster for the smallest sample size. While the running time of the spectral algorithm only increases polynomially, it increases exponentially for the BW algorithm. This can be seen by the straight lines for the running times of the BW algorithm on the $\mathrm{log}-\mathrm{log}$ scale. The differences in running time will therefore become even larger for increasing sample size.\\
Figure \ref{fig:compareBWa} shows the accuracy of the different learning instances. For three iterations, the accuracy of the BW algorithm with random initialization does not significantly increase for a larger sample size. Furthermore, the estimated parameters are not very accurate. For the BW algorithm initialized with learning results from the spectral algorithm, three iterations do not result in more accurate parameter estimates.\\
The spectral algorithm is significantly faster than the Baum-Welch algorithm. Already for small sample sizes, this advantage in speed is by many orders of magnitude. Only having been run for three iterations, the parameter estimates resulting from the BW algorithm are not more accurate than the results of the spectral algorithm. A further improvement of the solution would require many more iterations, further increasing the running time. A work by \cite{Kontorovich2013} addresses the convergence of the BW algorithm with respect to the number of iterations.\\

\section{Real-valued emissions}
\label{sec:real_valued_results}
In this section, the accuracy of parameter estimates obtained from sequences of real-valued emissions is evaluated. In contrast to the analysis in Section \ref{sec:comparison_algorithms}, sequences of real-valued emissions cannot be used as an input for the learning algorithms right away. It is therefore important to use a suitable method to prepare the data for the algorithms first.\\
One possible procedure, binning the emissions according to quantiles, was introduced in Section \ref{sec:real_valued_section}. In order to be able to quantify the accuracy of the parameter estimates learned from the data, synthetic sequences of real-valued observations were used.

\subsection{Generative process for synthetic experimental data} 
\begin{figure}
	\centering\includegraphics[scale=0.6]{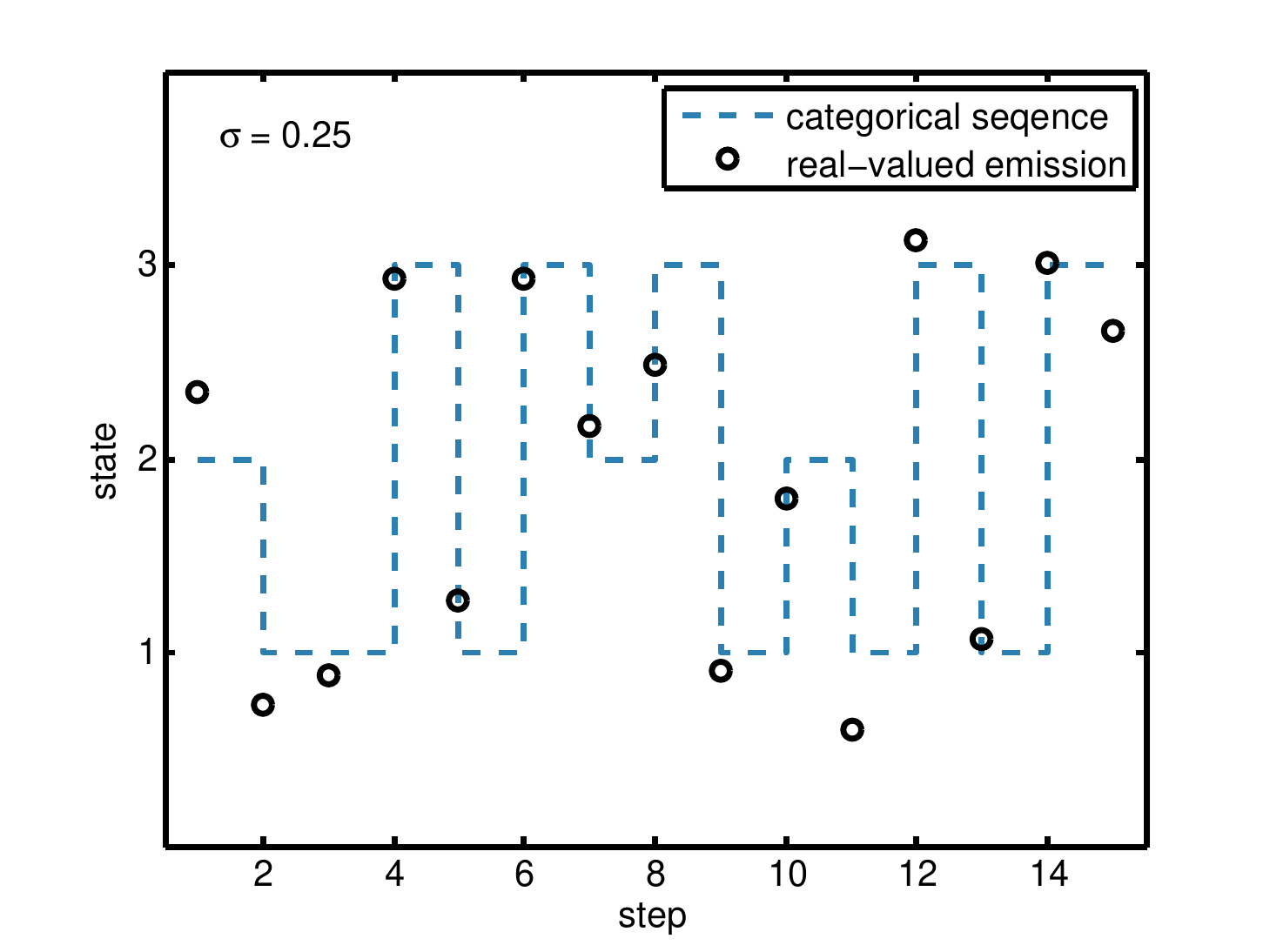}
	\caption{The generative process for a sequences of real-valued emissions. The categorical emissions (blue dashed line) were used as the mean values to generate a sequence of normally distributed random numbers with fixed standard deviation (here $\sigma = 0.25$).}
	\label{fig:generative_process}
\end{figure}

Sequences of categorical emissions were generated from the hidden Markov models specified in the beginning of Section \ref{sec:comparison_algorithms}. Only the two-state HMMs were considered. These sequences were then used to generate sequences of real valued emissions, resulting in our model experimental data.\\
Let there be a sequence of categorical emissions. The emissions are then used as the means for the generation of normally distributed random numbers with a fixed standard deviation $\sigma$. This results in a sequence of real-valued emissions. This generative process is depicted in Figure \ref{fig:generative_process}.

Learning from these sequences should therefore result in parameter estimates for the emission probability matrices and state transition matrices of the HMMs considered in this experiment. 

\subsection{Simple binning approach}
\label{subsec:simple_binning}
The model experimental data used to evaluate the simple binning approach was generated as specified by the generative process above. The model two-state HMMs ($k=2$, $d=3$ and $k=2$, $d=6$) are given in the beginning of Section \ref{sec:comparison_algorithms}.
Series of emissions generated using two different standard deviations, $\sigma = 0.1$ and $\sigma = 0.25$, were considered.\\
The following procedure was used to assess the simple binning approach:

\begin{enumerate}
	\item Given a sequence of real-valued emissions, use \textsc{SimpleBinning} to bin the emissions, creating a sequence of categorical emissions.
	\item Use this sequence of categorical emissions for learning the HMM model parameters with a spectral algorithm.
	\item Compare the resulting estimate of the emission probability matrix $\hat{O}$ and state transition probability matrix $\hat{T}$ to the known true values.
\end{enumerate}
Only the algorithm \textsc{learnAHK} was used for this analysis, since it was established in Section \ref{sec:comparison_algorithms} that this algorithm is the more accurate one of the two algorithms considered in this work.\\

\begin{figure}
        \centering
        \begin{subfigure}[b]{\textwidth}
                \centering\includegraphics[scale=0.8]{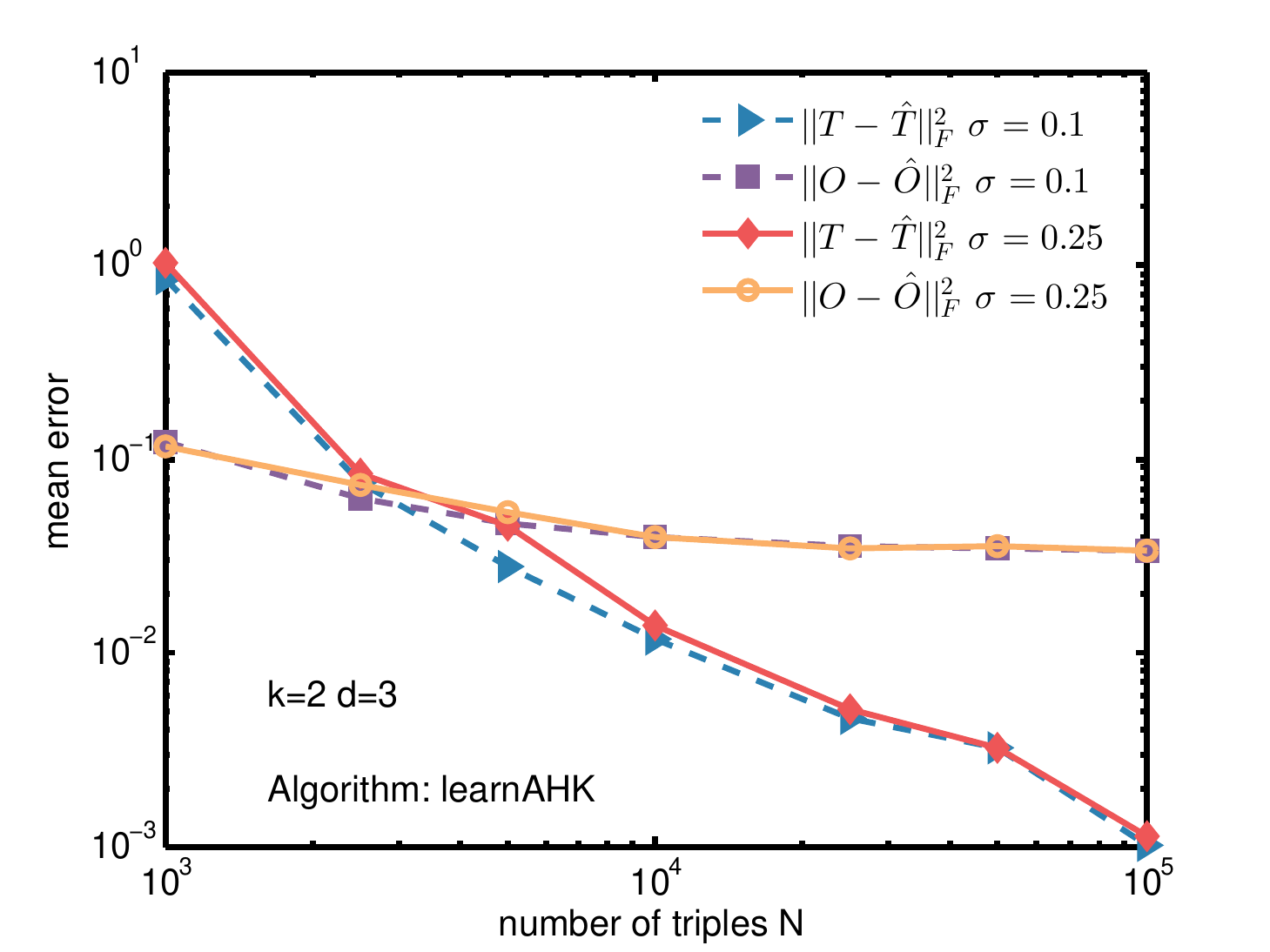}
                \caption{$k=2, d=3$}
								\label{fig:binning_simple_a}
        \end{subfigure}%
				\\
        \begin{subfigure}[b]{\textwidth}
                \centering\includegraphics[scale=0.8]{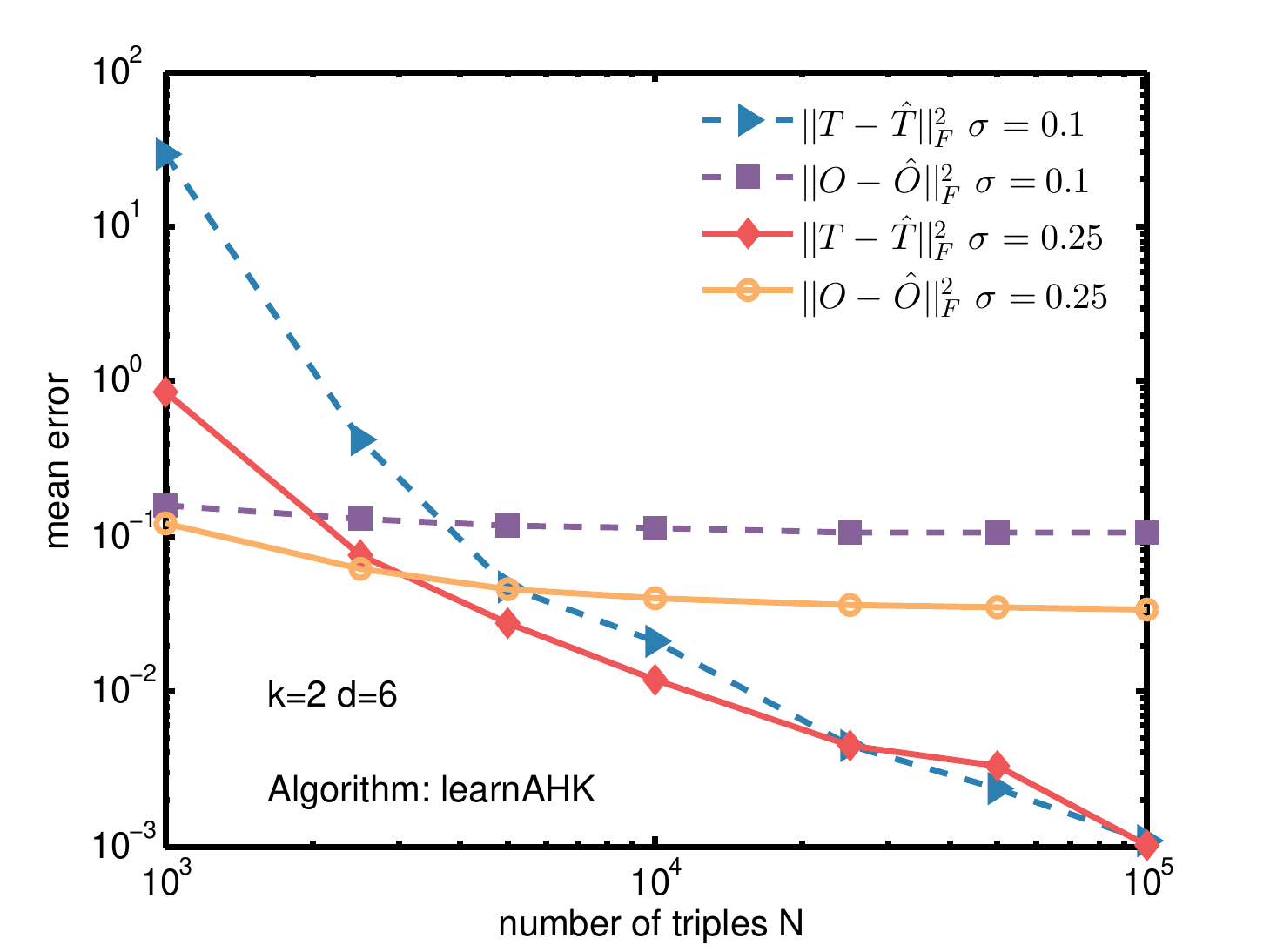}
                \caption{$k=2, d=6$}
                \label{fig:binning_simple_b}
        \end{subfigure}
        \caption{Accuracy of the parameter estimates using the simple binning approach. Two-state HMMs with three and six possible emissions, respectively.}
				\label{fig:binning_simple}
\end{figure}

The results for the simple binning approach are shown in Figure \ref{fig:binning_simple}. The accuracy of the parameter estimates as obtained by the procedure specified above was compared against the sample size.\\
For learning estimates of the state transition probability matrix $\hat{T}$, both datasets result in an equally accurate learning, with the estimation error decreasing with increasing sample size. This corresponds to the same behavior as observed for the categorical case. 



Considering the case for an estimate of the emission probability matrix $\hat{O}$, we see a different picture. Here an increased sample size does not result in a significant decrease in the estimation error. We can compare this to the results in Section \ref{sec:comparison_algorithms}, where the emission probability matrix was properly learned for categorical emissions. The real-valued emissions used in this analysis were based on the same model parameters. The inability to learn the estimate $\hat{O}$ is very likely to be due to the problem associated with equally sized bins when calculating the quantiles, as discussed in \ref{subsec:simple_binning_theory}.\\
The problems associated with the simple binning approach should not lead to the idea of binning by quantiles to be discarded though.

\subsection{Alternative approaches}
One possible solution to the problems associated with the simple binning approach is to bin on a finer scale. Instead of just binning into as many bins as the expected dimension of the emission probability matrix, one can consider using a larger number of bins.\\
Each time a real-valued observation is binned, information gets lost. The strength of the spectral methods lies partly in the fact that they can readily handle high observation spaces. This can be utilized with respect to the number of bins. When considering all sequences of real-valued emissions available, a larger number of quantiles can be calculated, resulting in a larger number of bins. The binning should ideally be finer than the smallest probability to be expected for a possible emission. Binning on a finer scale and retaining the information about the bin bounds leads to less information being lost.\\
When learning from time series binned on a finer scale, the spectral algorithms will return emission probability matrices of larger dimension than initially desired. Together with the bin bounds, these matrices can be used to recover probability distribution functions. \\
The result would be a mixture of probability distribution functions. Learning from mixtures of distributions is an active field of research and beyond the scope of this work. Among various available learning methods, spectral algorithms have also been developed for mixtures of distributions, see for example \cite{Achlioptas2005} and \cite{Dasgupta}.\\
The algorithm by \cite{Hsu2012} is only formulated for sequences of discrete observations. The algorithm by \cite{AHK2012}, however, also allows multivariate continuous observations. For this algorithm, categorical observations are represented by a basis vector in the $d$-dimensional standard coordinate system (see \ref{subsec:bagofwords}). It would also be possible to use a sequence of $d$-dimensional probability vectors as an input for this algorithm.\\
Instead of binning real-valued observations, where the challenge lies in finding appropriate bin bounds, one could therefore also try to find a representation of a scalar observation by a $d$-dimensional probability vector. The resulting sequence of vectors could then be used as the input for the learning algorithm \textsc{learnAHK}.


\chapter{Conclusion}
\label{cha:conclusion}
Spectral methods for learning hidden Markov models provide an appealing alternative to existing learning algorithms based on local search heuristics. The computational and sample complexity for existing search heuristics is oftentimes exponential, while it is only polynomial for the spectral methods considered in this work. Furthermore, spectral methods provide global estimates and are not prone to local optima. Both advantages become especially important when working with large observation spaces and large numbers of samples.\\
The synthetic observation sequences used in this work allowed to compare the obtained parameter estimates for example HMMs to their true values. It was established that the algorithm by \cite{AHK2012} provides better learning results than the one proposed by \cite{Hsu2012}. Both algorithms handle large observation spaces computationally efficiently. \\
For the state transition probability matrices, only the algorithm by \cite{AHK2012} learned reasonable parameter estimates for all four example HMMs considered. The algorithm by \cite{Hsu2012} did not obtain reasonable estimates for two of the example HMMs. It should be noted again, however, that the algorithm by \cite{Hsu2012} in its original form does not provide estimates for the HMM parameter matrices, and an additional computation step had to be implemented. The problems encountered for some of the example HMMs using the algorithm \textsc{learnHKZ} are consistent with instability concerns regarding this additional computation step by the authors in the original work.\\
Both algorithms failed to learn estimates for the emission probability matrices for two of the example HMMs, even for a large number of samples. These two HMMs were the ones with the largest observation spaces. Circumstances that are likely to have contributed to this are: instabilities encountered in numerical singular value and eigenvalue decompositions, the scaling of columns, and learned probabilities outside the interval $\left[ 0,1\right]$.\\
Comparing the algorithm by \cite{AHK2012} to the Baum-Welch algorithm, we found advantages in running time and accuracy of estimated quantities. Already for small sample sizes, the spectral algorithm was several orders of magnitude faster than the iterative method. The computational and sample complexity of polynomial order lets this advantage become even greater for larger sample sizes. This is even more significant when considering that the Baum-Welch algorithm was only run for few iterations. More iterations, which are generally needed for more accurate results, would increase the running time even further. Together with the spectral methods being free of local optima, this highlights major advantages over iterative methods. \\
The representation of sequences of real-valued observations to be used in the spectral algorithms proved to be challenging. The initial simple binning approach turned out to be not suitable for learning an estimate for the emission probability matrix. The state transition probability matrix was learned comparably well. This simple approach did not make full use of the advantages provided by the spectral methods though. The proposed idea to use a finer binning is only one possible approach for a better representation of real-valued emissions. However, this approach would require additional work after parameter estimates are obtained with the spectral algorithms.\\ \\
\textbf{Outlook}\\ \\
Spectral algorithms for learning hidden Markov models emerged only rather recently, and the two algorithms presented in this report are important examples of this approach. The tensor decomposition proposed by \cite{AGHKT2012} provides a method that might be less susceptible to instability. A possible future work would be to formulate an explicit learning algorithm for hidden Markov models based on this method, and compare it to the spectral methods considered in this work.\\
Regarding real-valued observations, further investigation is needed to find robust and reasonable representations for the use with spectral algorithms. Binning on a finer scale or representing real-valued observations by probability vectors are both two possible starting points for future considerations.

\appendix
\chapter{Appendix}
\label{cha:appendix}


\section{Proofs}

\subsubsection{Proposition \ref{prop_hmm}}
\begin{proof} Observing the conditional independence and using Bayes' theorem we find that
\begin{align*}
	\Pr \left[ h_1 = i | h_2 = j \right] &= \frac{\Pr \left[ h_2 = j | h_1 = i \right] \cdot \Pr \left[ h_1 = i\right]}{\Pr \left[ h_2 = j\right]} \\
		&= \frac{T_{ji} \pi_i}{\left( T \vec{\pi}\right)_j} 
			= \vec{e}_i \, \mathrm{diag} \left( \vec{\pi }\right) \, T^\top \, \mathrm{diag} \left( T \vec{\pi }\right)^{-1} \vec{e}_j.
\end{align*}
With definition \ref{def_general_setting} we therefore obtain
\[
	M_1 \vec{e}_j = \EE \left[ \vec{x}_1 | h_2 = j \right] = O \EE \left[ \vec{e}_{h_1} | h_2 = j \right] 
		= O \, \text{diag} \left( \vec{\pi }\right) \, T^\top \, \text{diag} \left( T \vec{\pi }\right)^{-1} \vec{e}_j.
\]
With the definitions of the observation probability matrix $O$ and the state transition probability matrix $T$ from section \ref{sec:hmmintro} we find
	\[
		M_2  \vec{e}_j = \EE \left[ \vec{x}_2 | h_2 = j \right] = O \vec{e}_j
	\]
	and
	\[
		M_3  \vec{e}_j = \EE \left[ \vec{x}_3 | h_2 = j \right] = O \EE \left[ \vec{e}_{h_3} | h_2 = j \right] = O T \vec{e}_j.
	\]
\end{proof}

\subsubsection{Lemma \ref{lem:secondthirdmoments}}
\begin{proof} Observing conditional independence, we have
\begin{align*}
 P_{3,1} &= \EE \left[ \EE \left[ \vec{x}_3 \otimes \vec{x}_1 | h \right] \right] \\
				 &= \EE \left[ \EE \left[ \vec{x}_3 | h \right] \otimes \EE \left[ \vec{x}_1 | h \right] \right] \\
				 &= \EE \left[ (M_3 \vec{e}_h) \otimes (M_1 \vec{e}_h) \right] \\
				 &= M_3 \left( \sum_{t=1}^k w_t \vec{e}_t \otimes \vec{e}_t \right) M_1^\top \\
				 &= M_3 \mathrm{diag}(\vec{w}) M_1^\top
\end{align*}
and
\begin{align*}
 P_{3,1,2} &= \EE \left[ \EE \left[ \left( \vec{x}_3 \otimes \vec{x}_1 \right) 
							\left\langle \vec{\eta},\vec{x}_2 \right\rangle | h\right] \right] \\
					&= \EE \left[ \EE \left[ \vec{x}_3 | h \right] \otimes \EE \left[ \vec{x}_1 | h \right]
					\left\langle \vec{\eta}, \EE \left[ \vec{x}_2|h \right] \right\rangle  \right] \\
					&= \EE \left[ (M_3 \vec{e}_h) \otimes (M_1 \vec{e}_h) 
					\left\langle \vec{\eta}, M_2 \vec{e}_h \right\rangle  \right] \\
					&= M_3 \left( \sum_{t=1}^k w_t \vec{e}_t \otimes \vec{e}_t 
					\left\langle \vec{\eta}, M_2 \vec{e}_h \right\rangle \right) M_1^\top \\
					&= M_3 \, \mathrm{diag}(M_2^\top \vec{\eta}) \, \mathrm{diag}(\vec{w}) \, M_1^\top.
\end{align*}
\end{proof}

\subsubsection{Lemma \ref{lem:observableoperator}}
\begin{proof}
Using Lemma \ref{lem:secondthirdmoments},
	\[
		U_3^\top P_{3,1}(\vec{\eta}) U_1 = \left( U_3^\top M_3 \right) \mathrm{diag}(\vec{w}) \left( M_1^\top U_1 \right)
	\]
is invertible by the assumption on $U_t$ and Condition \ref{condition1}. By using the relations from Lemma \ref{lem:secondthirdmoments} we obtain
	\begin{align*}
		B_{3,1,2}(\vec{\eta}) &= \left(U_3^\top P_{3,1,2}(\vec{\eta}) U_1 \right) \left(U_3^\top P_{3,1} U_1 \right)^{-1} \\
			& = \left(U_3^\top M_3 \, \mathrm{diag}(M_2^\top \vec{\eta}) \, \mathrm{diag}(\vec{w}) \, M_1^\top U_1 \right)
				\left(U_3^\top P_{3,1} U_1 \right)^{-1} \\
			& = \left(U_3^\top M_3 \right) \mathrm{diag}(M_2^\top \vec{\eta}) \left(U_3^\top M_3 \right)^{-1} \left(U_3^\top M_3 \right)
				\mathrm{diag}(\vec{w}) \left( M_1^\top U_1 \right) \left(U_3^\top P_{3,1} U_1 \right)^{-1} \\
			& = \left(U_3^\top M_3 \right) \mathrm{diag}(M_2^\top \vec{\eta}) \left(U_3^\top M_3 \right)^{-1}
				\left(U_3^\top P_{3,1} U_1 \right) \left(U_3^\top P_{3,1} U_1 \right)^{-1} \\
			&= \left(U_3^\top M_3 \right) \mathrm{diag}(M_2^\top \vec{\eta}) \left(U_3^\top M_3 \right)^{-1}.
	\end{align*}
\end{proof}

\subsubsection{Lemma \ref{lem:paramrecovery}}
\begin{proof}
From Equation \eqref{eq:empiricaloperator} and by setting $\vec{\eta} = U_2 \vec{\theta}_i$ we find $\forall i \in \left[ k \right]$
\begin{align*}
	\left(U_3^\top M_3 \right)^{-1} B_{3,1,2}(U_2 \vec{\theta}_i) \left(U_3^\top M_3 \right) 
		&= \mathrm{diag} (M_2^\top U_2 \vec{\theta}_i) \\
		&= \mathrm{diag} ( \langle \vec{\theta}_i, U_2^\top M_2 \vec{e}_1 \rangle, \langle \vec{\theta}_i, U_2^\top M_2 \vec{e}_2 \rangle,
			\ldots \langle \vec{\theta}_i, U_2^\top M_2 \vec{e}_k \rangle \\
		&= \mathrm{diag} \left( \lambda_{i,1}, \lambda_{i,2}, \ldots, \lambda_{i,k} \right).
\end{align*}
Using the definition of $\Theta \in \RR^{k \times k}$, we obtain for $L \in \RR^{k \times k}$
	\begin{align*}
		L &= 
			\begin{pmatrix}
				\lambda_{1,1} & \lambda_{1,2} & \cdots & \lambda_{1,k} \\
				\lambda_{2,1} & \lambda_{2,2} & \cdots & \lambda_{2,k} \\
				\vdots & \vdots & \ddots & \vdots \\
				\lambda_{k,1} & \lambda_{k,2} & \cdots & \lambda_{k,k}
			\end{pmatrix} \\
			&= 
			\begin{pmatrix}
				\langle \vec{\theta}_1, U_2^\top M_2 \vec{e}_1 \rangle & \langle \vec{\theta}_1, U_2^\top M_2 \vec{e}_2 \rangle &
					\cdots & \langle \vec{\theta}_1, U_2^\top M_2 \vec{e}_k \rangle \\
				\langle \vec{\theta}_2, U_2^\top M_2 \vec{e}_1 \rangle & \langle \vec{\theta}_2, U_2^\top M_2 \vec{e}_2 \rangle &
					\cdots & \langle \vec{\theta}_2, U_2^\top M_2 \vec{e}_k \rangle \\	
				\vdots & \vdots & \ddots & \vdots \\
					\langle \vec{\theta}_k, U_2^\top M_2 \vec{e}_1 \rangle & \langle \vec{\theta}_k, U_2^\top M_2 \vec{e}_2 \rangle &
					\cdots & \langle \vec{\theta}_k, U_2^\top M_2 \vec{e}_k \rangle
			\end{pmatrix} \\
			&= \Theta U_2^\top M_2.
	\end{align*}
\end{proof}

\section{Specifications of computing environment}
\label{sec:appendix_specs}
Computations were conducted on a personal computer. \\ \\
\textbf{Relevant specifications:}
\begin{itemize}
	\item Intel Core i5-540M
	\item 4.00 GB RAM
	\item 160 GB Intel solid state drive
	\item Microsoft Windows 8.1 Professional 64-bit
	\item MathWorks Matlab R2012b 64-bit
\end{itemize}

\section{Source code}
\label{sec:source_code}
The algorithms \textsc{learnHKZ} and \textsc{learnAHK} were implemented in Matlab. The source code, along with the binning routine used in this work, is available at \url{https://github.com/cmgithub/spectral}.

\backmatter

\bibliographystyle{ThesisStyleWithEtAl}
\bibliography{library}

\end{document}